\documentclass[journal]{IEEEtran}
\usepackage{amssymb,amsfonts,amsmath,amscd}
\usepackage[ruled,vlined,linesnumbered]{algorithm2e}  %
\usepackage{array}
\usepackage[caption=false,font=footnotesize,labelfont=rm,textfont=rm]{subfig}
\usepackage{textcomp}
\usepackage{stfloats}
\usepackage{url}
\usepackage{verbatim}
\usepackage{graphicx}
\usepackage{subfig}
\usepackage{cite}
\usepackage{hyperref}

\usepackage{multirow}
\usepackage{makecell}
\usepackage{bm}
\hypersetup{
colorlinks=true,
allcolors=blue,
linkcolor=blue,
anchorcolor=blue,
citecolor=blue}
\usepackage{threeparttable}
\usepackage{booktabs}

\DeclareMathAlphabet{\mathscr}{OT1}{pzc}{m}{it}

\begin{document}
%
\title{PaddingFlow: Improving Normalizing Flows with Padding-Dimensional Noise}
%
%

\author{Qinglong~Meng,
        Chongkun Xia,
        Xueqian Wang*
        \thanks{* Corresponding author}
        \thanks{
        This work was supported by the National Key R\&D Program of China (2022YFB4701400/4701402), 
        National Natural Science Foundation of China (No. U21B6002, 62203260), 
        Guangdong Basic and Applied Basic Research Foundation (2023A1515011773).}
        \thanks{
        Qinglong Meng, and Xueqian Wang are with Tsinghua Shenzhen International Graduate School, Shenzhen 518055, China (e-mail: mengql22@mails.tsinghua.edu.cn; wang.xq@sz.tsinghua.edu.cn;).}
        \thanks{
        Chongkun Xia is with School of Advanced Manufacturing, Sun Yat-sen University, Shenzhen 518107, China (e-mail: xiachk5@mail.sysu.edu.cn;).}}

%
%

\markboth{Journal of \LaTeX\ Class Files,~Vol.~14, No.~8, August~2015}%
{Shell \MakeLowercase{\textit{et al.}}: Bare Demo of IEEEtran.cls for IEEE Journals}
%



\maketitle
\begin{abstract}
Normalizing flow is a generative modeling approach with efficient sampling. However, Flow-based models suffer two issues: 1) If the target distribution is manifold, due to the unmatch between the dimensions of the latent target distribution and the data distribution, flow-based models might perform badly. 2) Discrete data might make flow-based models collapse into a degenerate mixture of point masses. To sidestep such two issues, we propose PaddingFlow, a novel dequantization method, which improves normalizing flows with padding-dimensional noise. To implement PaddingFlow, only the dimension of normalizing flows needs to be modified. Thus, our method is easy to implement and computationally cheap. Moreover, the padding-dimensional noise is only added to the padding dimension, which means PaddingFlow can dequantize without changing data distributions. Implementing existing dequantization methods needs to change data distributions, which might degrade performance. We validate our method on the main benchmarks of unconditional density estimation, including five tabular datasets and four image datasets for Variational Autoencoder (VAE) models, and the Inverse Kinematics (IK) experiments which are conditional density estimation. The results show that PaddingFlow can perform better in all experiments in this paper, which means PaddingFlow is widely suitable for various tasks. The code is available at: \href{https://github.com/AdamQLMeng/PaddingFlow}{https://github.com/AdamQLMeng/PaddingFlow}.
\end{abstract}


%
\IEEEpeerreviewmaketitle

\section{Introduction}\label{sec.intro}
\IEEEPARstart{N}{omalizing} flow (NF) is one of the widely used generative modeling approaches. Flow-based generative models use cheaply invertible neural networks and are easy to sample from. However, two issues limit the performance of flow-based generative models: 1) Mismatch of the latent target distribution dimension and the data distribution dimension \cite{softflow}; 2) Discrete data leads normalizing flows to collapse to a degenerate mixture of point masses \cite{discrete}. Here, we list five key features that an ideal dequantization for sidestepping such two issues should offer:
\begin{itemize}\label{sec.5charac}
\vspace{2pt}
	\item\textbf{1. Easy to implement:} To implement dequantization, the modification of the original models should be simple.
\vspace{2pt}
    \item\textbf{2. Not have to change the data distribution:} Changing the data distribution may bring benefits because of the reasonable assumption of the neighbor of the data points. On the other side, any assumption may be unreasonable sometimes. Dequantization should not change the data distribution if it is needed.
\vspace{2pt}
    \item\textbf{3. Unbiased estimation:} The generative samples should be the unbiased estimations of the data.
\vspace{2pt}
    \item\textbf{4. Not computationally expensive:} At first, dequantization methods as the preprocessing will degrade the inference speed if the computation is large. Secondly, the improvement provided by dequantization methods is limited sometimes, large computations will make the method poor economic.
\vspace{2pt}
    \item\textbf{5. Widely suitable:} If the selection of dequantization methods is complicated, using dequantization methods is poor economic. Thus, dequantization should provide improvement for various tasks.
\vspace{2pt}
\end{itemize}
\par Prior work has proposed several dequantization methods, including uniform dequantization, variational quantization \cite{flowpp}, and conditional quantization \cite{softflow}. Uniform dequantization is easy to implement and computationally cheap, however, the generative samples are biased estimations (Eq. \ref{eq.exp_uni}). Variational dequantization can provide improvement to various tasks, but using an extra flow-based model to generate noise is computationally expensive, and difficult to implement because there is an extra model that needs to be trained. Conditional dequantization is to add a conditional distribution where the condition is the variance of noise sampled from a uniform distribution. Such a complicated distribution might degrade performance significantly (Fig. \ref{fig.toy}). Moreover, it requires modifying the original models to conditional normalizing flows, which can be intractable when the original model is an unconditional normalizing flow.
\par In this paper, we propose a novel dequantization method that can satisfy the five key features we list, named PaddingFlow, which improves normalizing flows with padding-dimensional noise. Unlike all prior work, PaddingFlow can dequantize without changing the data distribution. To implement PaddingFlow, only the dimension of distribution needs to be modified. Thus, unlike variational dequantization, the computation of implementing PaddingFlow is relatively low. We validate our method on 9 density estimation benchmarks (including 5 tabular datasets, and 4 VAE datasets), and IK experiments, which contain both unconditional and conditional density estimation. The results show that PaddingFlow can perform better and is suitable for both discrete and continuous normalizing flows.

\begin{figure*}[t]
	\centering
	\subfloat{
		\includegraphics[width=0.95\textwidth]{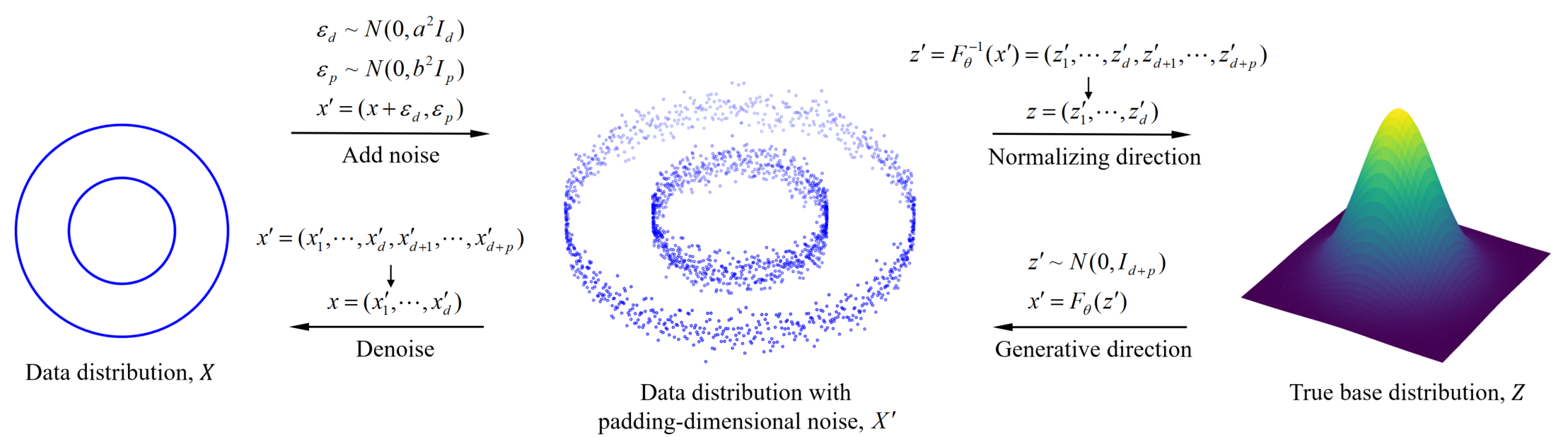}}
    \caption{Implementation of PaddingFlow for training flow-based models. $d$ denotes the dimension of the data distribution, and $p$ denotes the dimension of padding-dimensional noise. }
    \label{fig.method}
\end{figure*}

\section{Background}\label{sec.bg}
In this section, we introduce Normalizing Flows (NFs) briefly. Moreover, the prior work for sidestepping the two issues abovementioned are introduced, and the main flaws of these methods are analyzed.
\subsection{Normalizing Flows}\label{sec.nf}
Normalizing flows are to model the target distribution as a transformation $F_{\theta}$ of the base distribution, which is usually Gaussian distribution:
\begin{equation}
x=F_{\theta}(z), \ \mathrm{where} \ z\sim N(0,I).\label{eq.trans}
\end{equation}
Furtherly, the density of $x$ can be obtained by a change of variables:
\begin{equation}
p_{X}(x)=p_{Z}(F_{\theta}^{-1}(x))|J_{F_{\theta}^{-1}}(x)|.\label{eq.density}
\end{equation}
\par In practice, we often construct a neural network to fit the transformation $F_{\theta}$. The corresponding objective function is usually Kullback-Leibler (KL) divergence to minimize divergence between the flow-based model $p_{X}(x;\theta)$ and the target distribution $p_{X}^{*}(x)$ \cite{survey}, which can be written as:
\begin{equation}
\begin{aligned}
\mathcal{L}(x;\theta)&=D_{KL}[p_{X}^{*}(x)\Vert p_{X}(x;\theta)]\\
&=-\mathbb{E}_{p_{X}^{*}(x)}[\mathrm{log}p_{X}(x;\theta)]-H[p_{X}^{*}(x)]\\
&=-\mathbb{E}_{p_{X}^{*}(x)}[\mathrm{log}p_{Z}(F_{\theta}^{-1}(x))+\mathrm{log|det}J_{F_{\theta}^{-1}}(x)|]\\
&-H[p_{X}^{*}(x)].\label{eq.kl}
\end{aligned}
\end{equation}
Due to the data distribution $p_{X}^{*}(x)$ is fixed, the second term is a constant. And the expectation over $p_{X}^{*}(x)$ can be estimated by Monte Carlo. Therefore, the loss function can be written as:
\begin{equation}
\begin{aligned}
\mathcal{L}(\mathcal{X};\theta)\approx-\frac{1}{N}\sum_{i=1}^{N}[&\mathrm{log}p_{Z}(F_{\theta}^{-1}(^{(i)}x))\\
&+\mathrm{log|det}J_{F_{\theta}^{-1}}(^{(i)}x)|],\label{eq.loss}
\end{aligned}
\end{equation}
where $\mathcal{X}=\{x_{i}\}_{i=1}^{N}$. As for continuous normalizing flows (CNF)\cite{cnf}\cite{ffjord}, the computation of total change in log-density $\mathrm{log} p_{Z}(z)$ is done by integrating across time, and it is the integration of the trace of the Jacobian matrix, instead of using the determinant:
\begin{equation}
\mathrm{log} p_{Z}(z(t_{1}))=\mathrm{log} p_{Z}(z(t_{0}))-\int_{t_{0}}^{t_{1}}\mathrm{Tr} \left( J_{F_{\theta}}(z) \right),\label{eq.cnf}
\end{equation}
which simplifies the computation of the change of log-density.

\subsection{Dequantization}\label{sec.dequan}
\par The real-world datasets, such as MNIST\cite{mnist} and UCI datasets\cite{uci}, are recordings of continuous signals quantized into discrete representations. Training a flow-based model on such datasets is to fit a continuous density model to discrete distribution. Moreover, the latent target distribution in some tasks is manifold, such as some conditional distributions, which means the dimension of the target distribution is lower than the data dimension. Both two issues will hurt the training loss and generalization. To sidestep these issues, several dequantization methods were proposed. In this section, the representative work will be introduced, and the main flaws will be analyzed as well.
\par\textbf{Uniform Dequantization} is used most widely in prior work, due to the simple noise formula and no need to modify the model for adapting such a dequantization method. However, uniform noise will lead models to suboptimal solutions. Here, we give a simplified example for explanation. After adding uniform noise ($u\sim U(0,1)$) to the normalized data ($x\sim N(0,1)$), the density of the noisy data ($y$) can be written as:
\begin{equation}
\begin{aligned}
p_{Y}(y)&=\int_{-\infty}^{+\infty}p_{X}(y-u)p_{U}(u)du\\
&=\int_{0}^{1}\frac{1}{\sqrt{2\pi}}e^{-\frac{1}{2}(y-u)^{2}}du.
\end{aligned}
\end{equation}
We further compute the expectation of $Y$ as follows:
\begin{equation}
\mathbb{E}(Y)=\int_{-\infty}^{+\infty}ydy\int_{0}^{1}\frac{1}{\sqrt{2\pi}}e^{-\frac{1}{2}(y-u)^{2}}du=1-e^{-\frac{1}{2}}.\label{eq.exp_uni}\footnote[1]{The details of computation can be found in Appendix B.}
\end{equation}
Therefore, the data generated by the flow-based generative models trained on the data added uniform noise is a biased estimation of the original data. If the interval is symmetric (i.e. $u\sim U(-a, a)$), the expectation of $Y$: $\mathbb{E}(Y)=0$, which means the estimation is unbiased. However, another issue is that assigning uniform density to the unit hypercubes $x+[0,1)^{D}$ is difficult and unnatural for neural network density models.
\par\textbf{Variational dequantization\cite{flowpp}} is proposed to sidestep the issue that the uniform noise assigns uniform density to the unit hypercubes $x+[0,1)^{D}$. Variational dequantization uses an extra flow-based generative model to generate noise, which makes noise obey an arbitrary conditional distribution where the condition is the corresponding data ($x$). The process can be written as:
\begin{equation}
\left\{
\begin{array}{ll}
\epsilon'=g(\epsilon|x), \mathrm{where}\ \epsilon\sim N(0,I)&\\[2pt]
z=f^{-1}(x+\epsilon')&\\
\end{array}\right.\label{eq.flowpp}
\end{equation}
where $g(\cdot)$ is the dequantization flow, $f(\cdot)$ is the data flow. Apparently, such a method is computationally expensive especially when the data is images.
\par\textbf{Conditional dequantization\cite{softflow}} is to use a conditional distribution for generating noise as well as variational dequantization. Instead of using an extra model, SoftFlow\cite{softflow} uses the parameterization trick as VAE models\cite{vae}. The process can be written as:
\begin{equation}
\left\{
\begin{array}{ll}
\epsilon\sim N(0,I)&\\[2pt]
c\sim U(0,1)&\\[2pt]
z=f^{-1}(x+c\cdot\epsilon|c)&\\
\end{array}\right.\label{eq.softflow}
\end{equation}
In the generative direction, SoftFlow set the condition $c$ and the noise $\epsilon$ to $\vec{0}$ to remove the effect of the noise. This method requires the model to be conditional normalizing flow. Especially when the original model is an unconditional normalizing flow, the modification can be intractable. Additionally, conditional dequantization is to add an extra conditional distribution to the original distribution. As shown in Fig. \ref{fig.toy}, when the target distribution is also conditional, it might degrade performance significantly.
\par In conclusion, the prior work shows promising results but doesn't satisfy all key features we list (Sec. \ref{sec.5charac}). In this paper, we propose PaddingFlow, which improves normalizing flows with padding-dimensional noise. Our method satisfies all five key features we list. In particular, PaddingFlow can overcome the limitation that existing dequantization methods have to change the data distribution.

\begin{figure*}[t]
	\centering
	\subfloat{
		\includegraphics[width=0.95\textwidth]{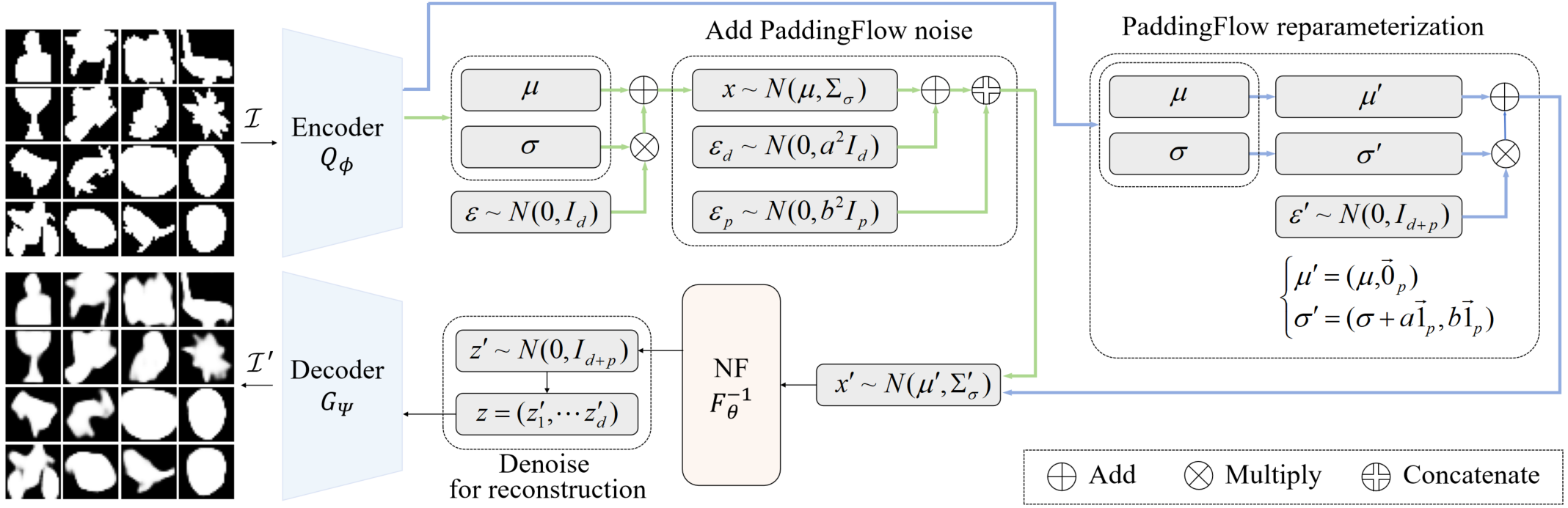}}
    \caption{Two ways of implementing PaddingFlow on a VAE model: 1) adding PaddingFlow noise (green lines), and 2) PaddingFlow reparameterization (blue lines). Images ($\mathcal{I}^{'}$) shown in the figure are reconstructed by the PaddingFlow-based VAE model trained on Caltech 101 Silhouettes.}
    \label{fig.vae}
\end{figure*}

\section{PaddingFlow}\label{sec.method}
In this section, we introduce the formula of PaddingFlow noise, and the implementation of plain normalizing flows and flow-based VAE models. Moreover, for VAE models, we also propose PaddingFlow parameterization. 
\subsection{PaddingFlow Noise}\label{sec:pd_noise}
To overcome the limitations of noise added to data directly, we proposed the padding-dimensional noise denoted as $\varepsilon_{p}$, which doesn't change the distribution of data dimensions. Moreover, to inherit the merits of uniform noise and achieve unbiased estimation, we choose to add the Gaussian noise with zero expectation $N(0,I)$ to data as a complement of padding-dimensional noise, denoted as $\varepsilon_{d}$. $\varepsilon_{d}$ is called data noise in this paper. Furthermore, the variances of $\varepsilon_{d}$ and $\varepsilon_{p}$ should vary depending on the density of data points and the scale of data respectively. We introduce hyperparameters of variances to the noise, which means the distributions are $N(0,a^{2}I)$, and $N(0,b^{2}I)$. Therefore, the formula of PaddingFlow noise can be written as (Fig. \ref{fig.method}):
\begin{equation}
\left\{
\begin{array}{ll}
\varepsilon_{d}\sim N(0,a^{2}I_{d}) &\\[2pt]
\varepsilon_{p}\sim N(0,b^{2}I_{p}) &\\[2pt]
x'= (x+\varepsilon_{d},\varepsilon_{p})&\\
\end{array}\right.\label{eq.add}
\end{equation}
where $a$, and $b$ denote the variances of data noise, and padding-dimensional noise respectively; $d$ and $p$ denote the dimension of data noise, and padding-dimensional noise respectively.
\par After implementing our method, in the normalizing direction, it only needs to cut out the first data dimensions of the normalized point ($z'$) for obtaining the point from true base distribution ($N(0, I_{d})$):
\begin{equation}
\left\{
\begin{array}{ll}
z'=F_{\theta}^{-1}(x')\sim N(0,I_{d+p}) &\\[2pt]
z=(z'_{1},\cdots,z'_{d})\sim N(0,I_{d}) &\\
\end{array}\right.\label{eq.norm}.
\end{equation}
\par In the generative direction, the operation of obtaining the true generative data is the same as the normalizing direction:
\begin{equation}
\left\{
\begin{array}{ll}
x'=F_{\theta}(z') &\\
x=(x'_{1},\cdots,x'_{d}) &\\
\end{array}\right.\label{eq.gen}.
\end{equation}
\par As for the loss function, it should be written as:
\begin{equation}
\begin{aligned}
\mathcal{L}(\mathcal{X}';\theta)\approx-\frac{1}{N}\sum_{i=1}^{N}[&\mathrm{log}p_{Z'}(F_{\theta}^{-1}(^{(i)}x'))\\
&+\mathrm{log|det}J_{F_{\theta}^{-1}}(^{(i)}x')|],\label{eq.ploss}
\end{aligned}
\end{equation}
where $\mathcal{X}'=\{x'_{i}\}_{i=1}^{N}$. 

\begin{figure*}[t]
  	\begin{minipage}{0.1\linewidth}
        \vspace{0.8\linewidth}
 		\centerline{\small\rotatebox[origin=c]{90}{Data}}
        \vspace{0.8\linewidth}
        \centerline{\small\rotatebox[origin=c]{90}{FFJORD}}
        \vspace{0.7\linewidth}
        \centerline{\small\rotatebox[origin=c]{90}{SoftFlow}}
        \vspace{0.55\linewidth}
        \centerline{\small\rotatebox[origin=c]{90}{PaddingFlow}}
	\end{minipage}
\hspace{-20pt}
 	\begin{minipage}{0.13\linewidth}
        \vspace{4pt}
 		\centerline{\makecell{circles}}
        \vspace{10.5pt}
        \includegraphics[width=\textwidth]{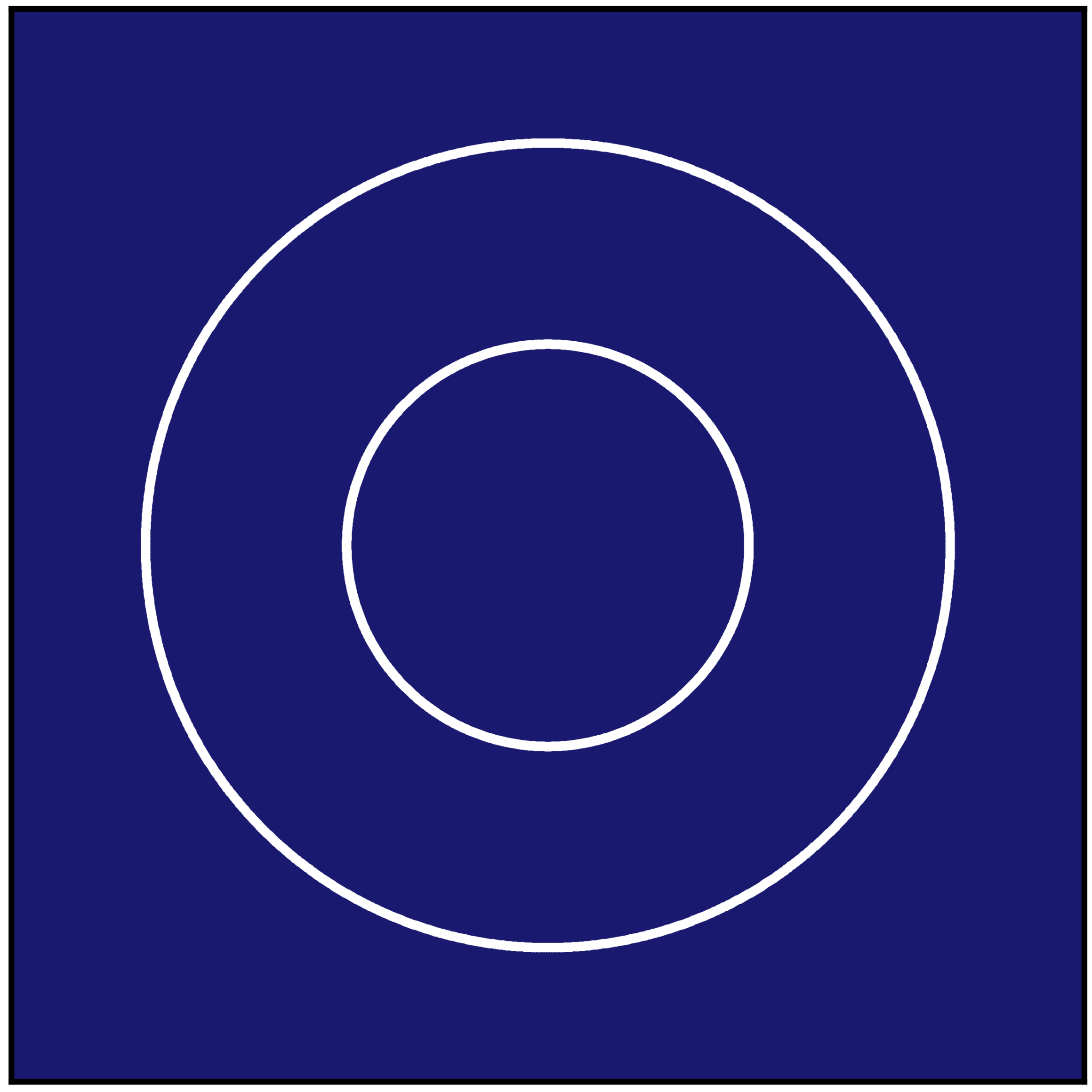}
        \vspace{2pt}
        \includegraphics[width=\textwidth]{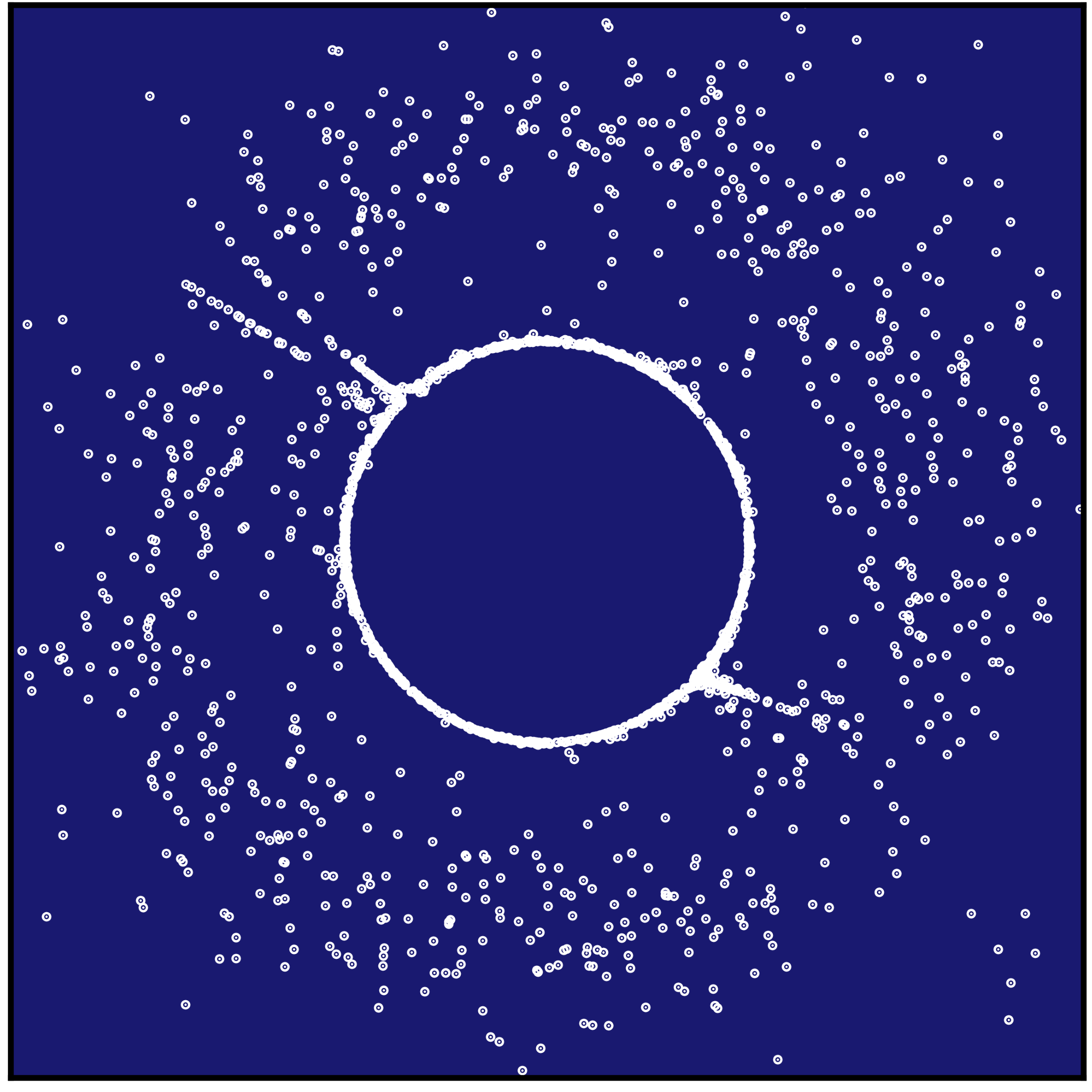}
        \vspace{2pt}
        \includegraphics[width=\textwidth]{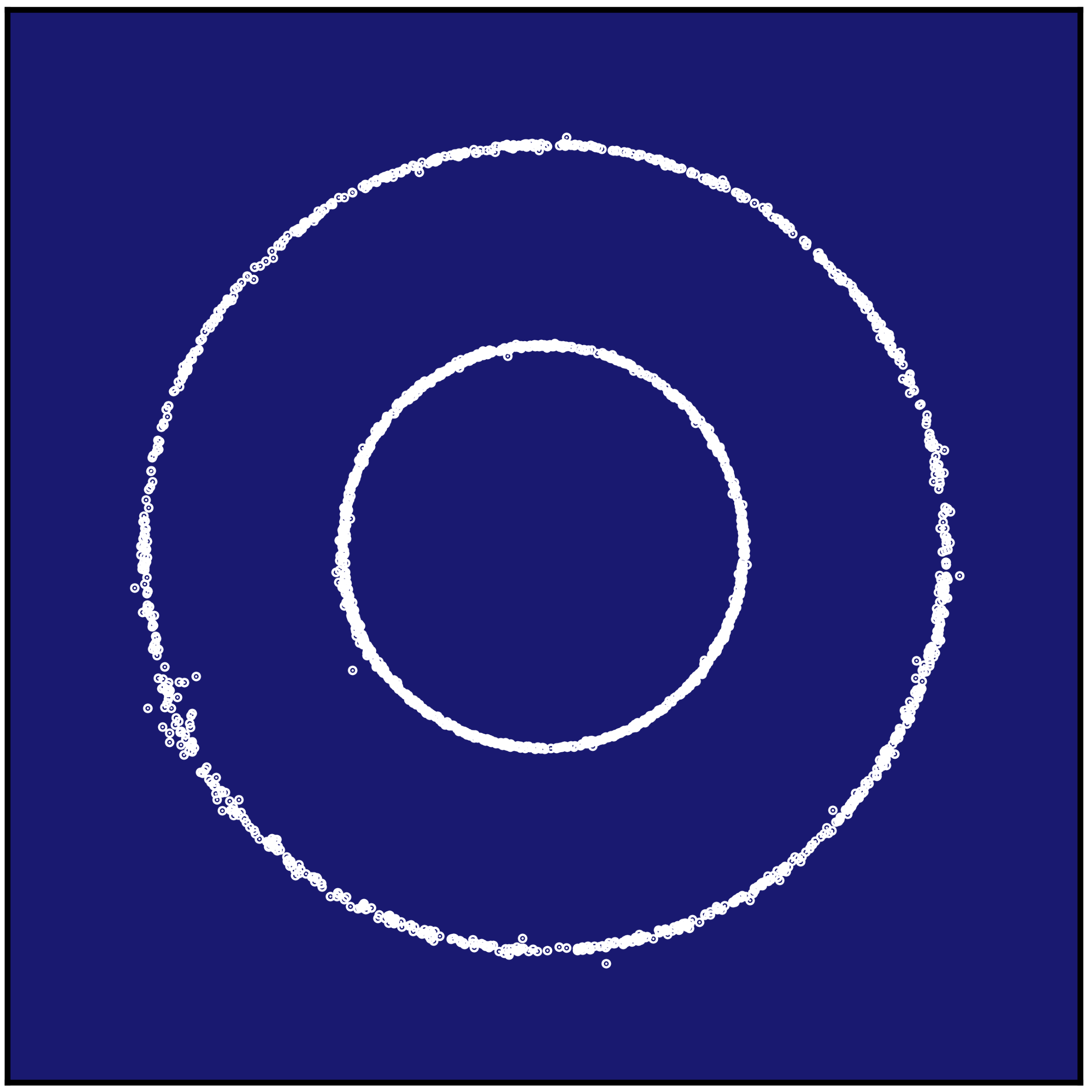}
        \vspace{2pt}
        \includegraphics[width=\textwidth]{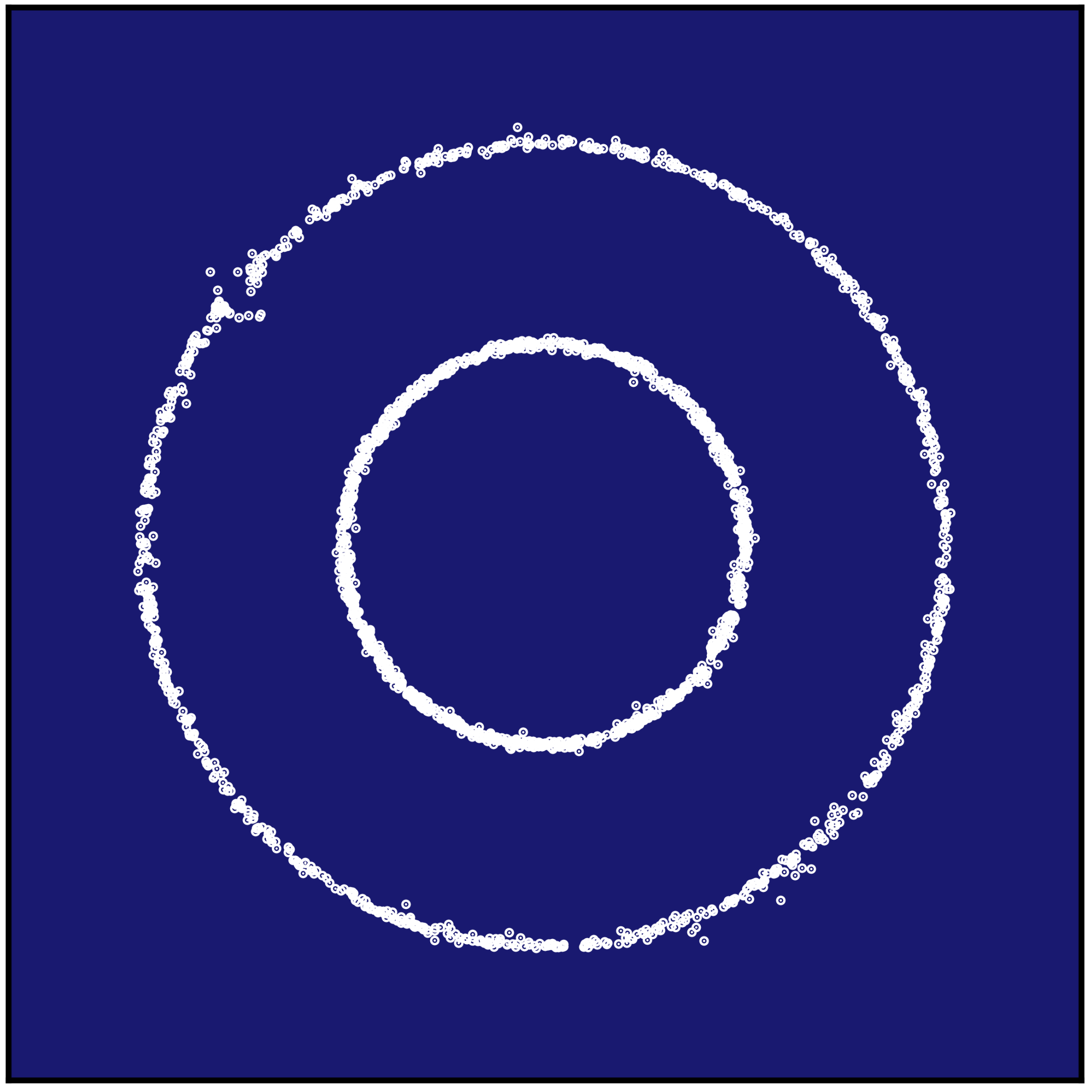}
	\end{minipage}
\hspace{2pt}
    \begin{minipage}{0.27\linewidth}
 		\centerline{\makecell{conditional circles\\\hspace{10pt}(c=0.5)\hspace{30pt}(c=0.25)}}
		\vspace{3pt}
        \subfloat{
        \includegraphics[width=0.48\textwidth]{figs/toy/circles0.5.png}
        \includegraphics[width=0.48\textwidth]{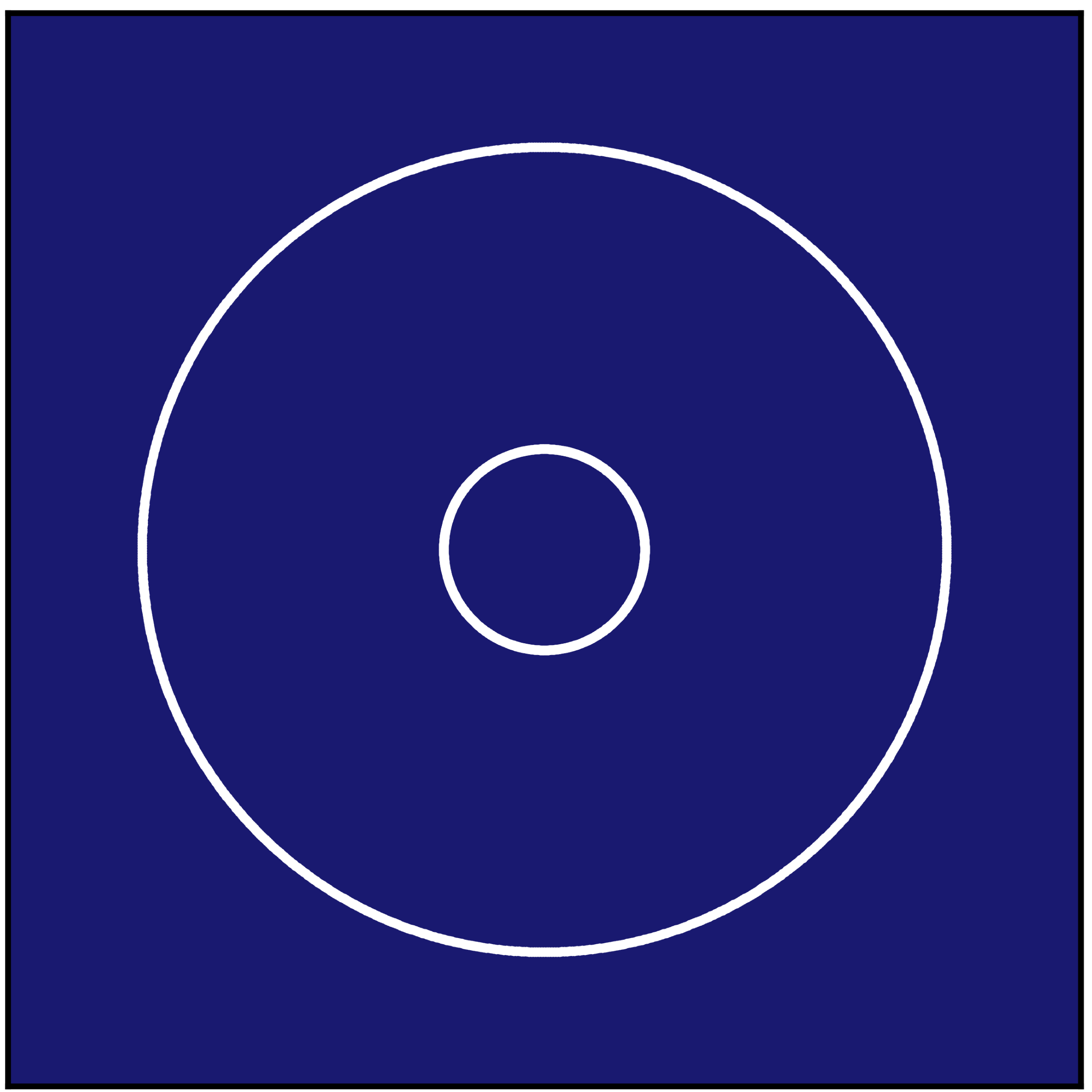}
        }\\[-8pt]
        \subfloat{
        \includegraphics[width=0.48\textwidth]{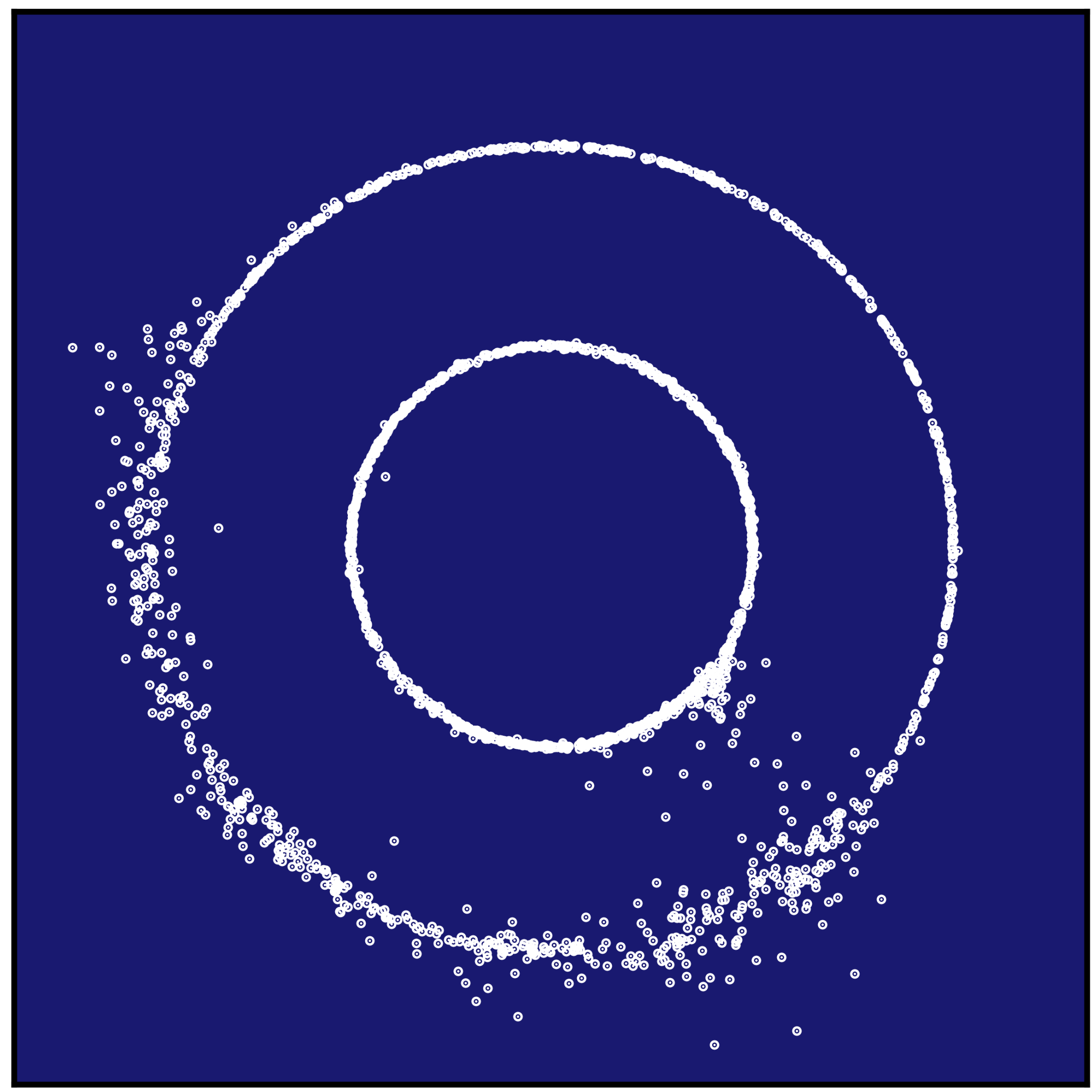}
        \includegraphics[width=0.48\textwidth]{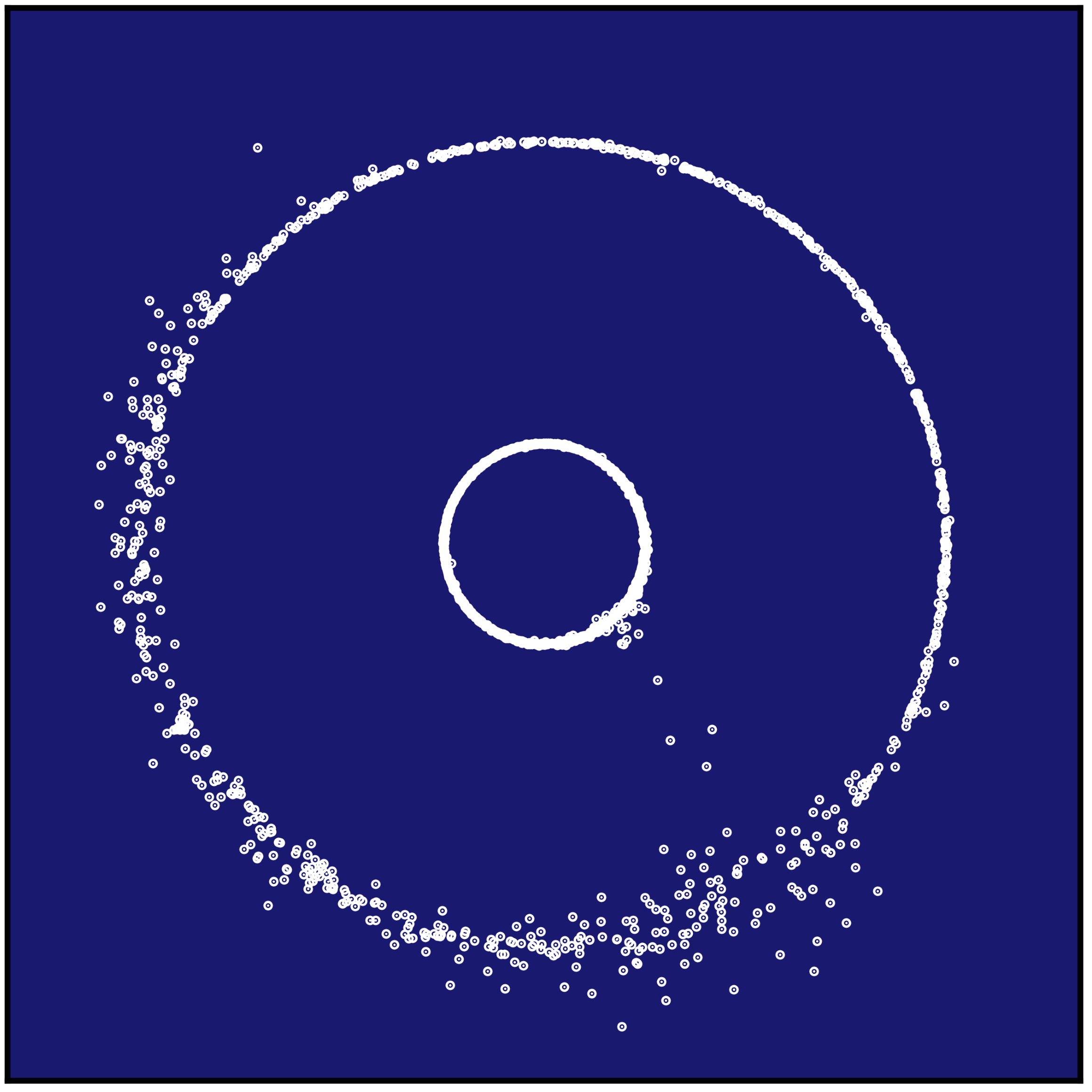}
        }\\[-8pt]
        \subfloat{
        \includegraphics[width=0.48\textwidth]{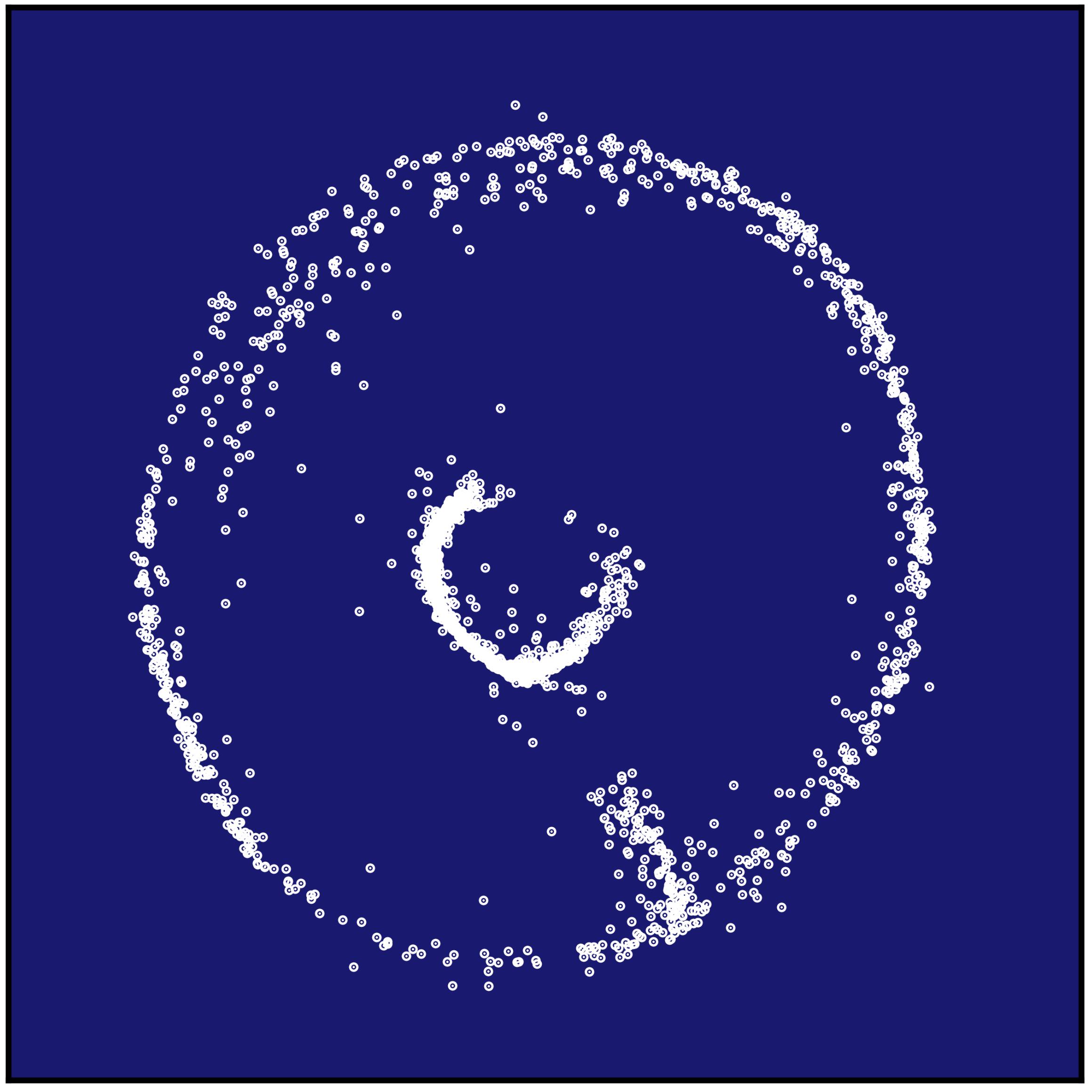}
        \includegraphics[width=0.48\textwidth]{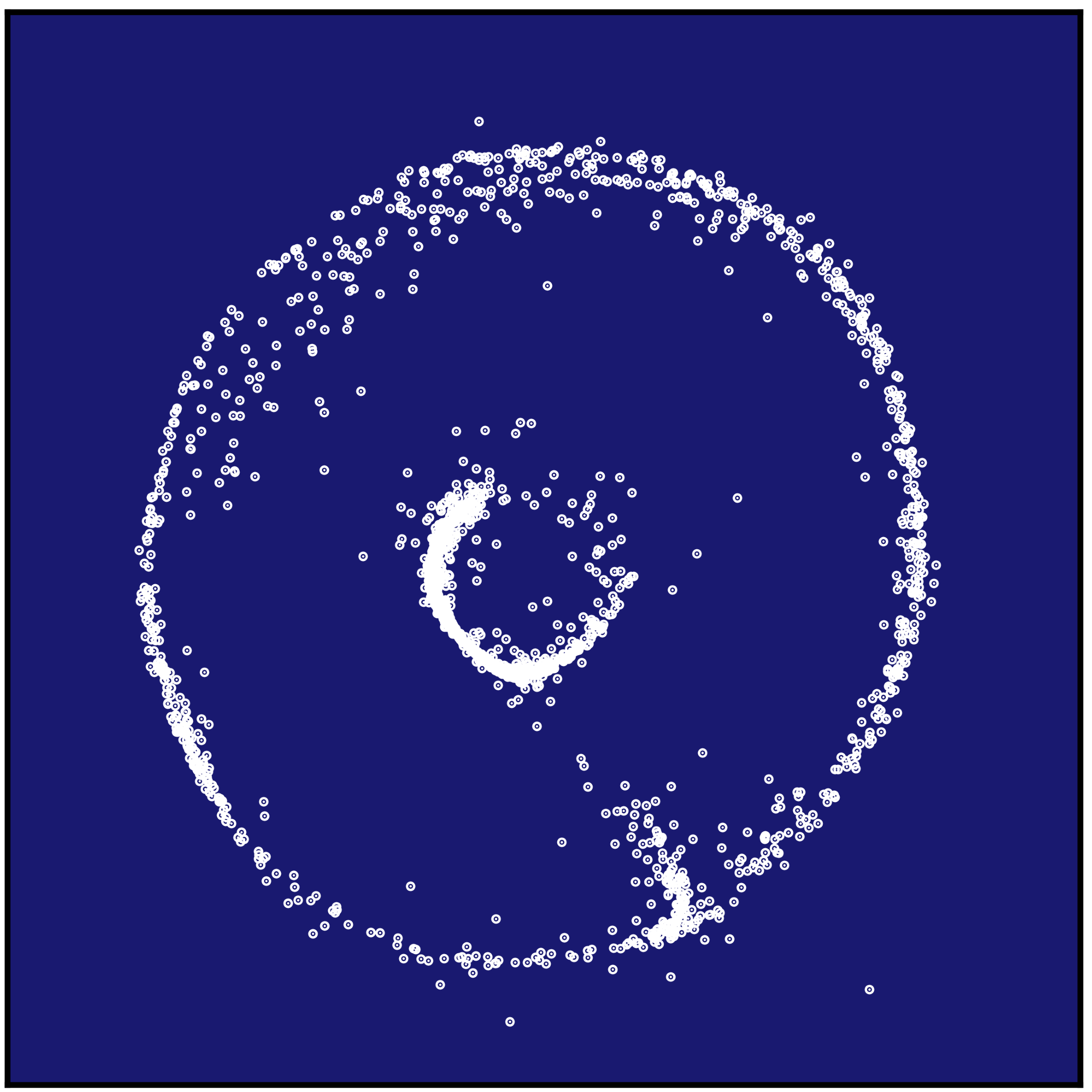}
        }\\[-8pt]
        \subfloat{
        \includegraphics[width=0.48\textwidth]{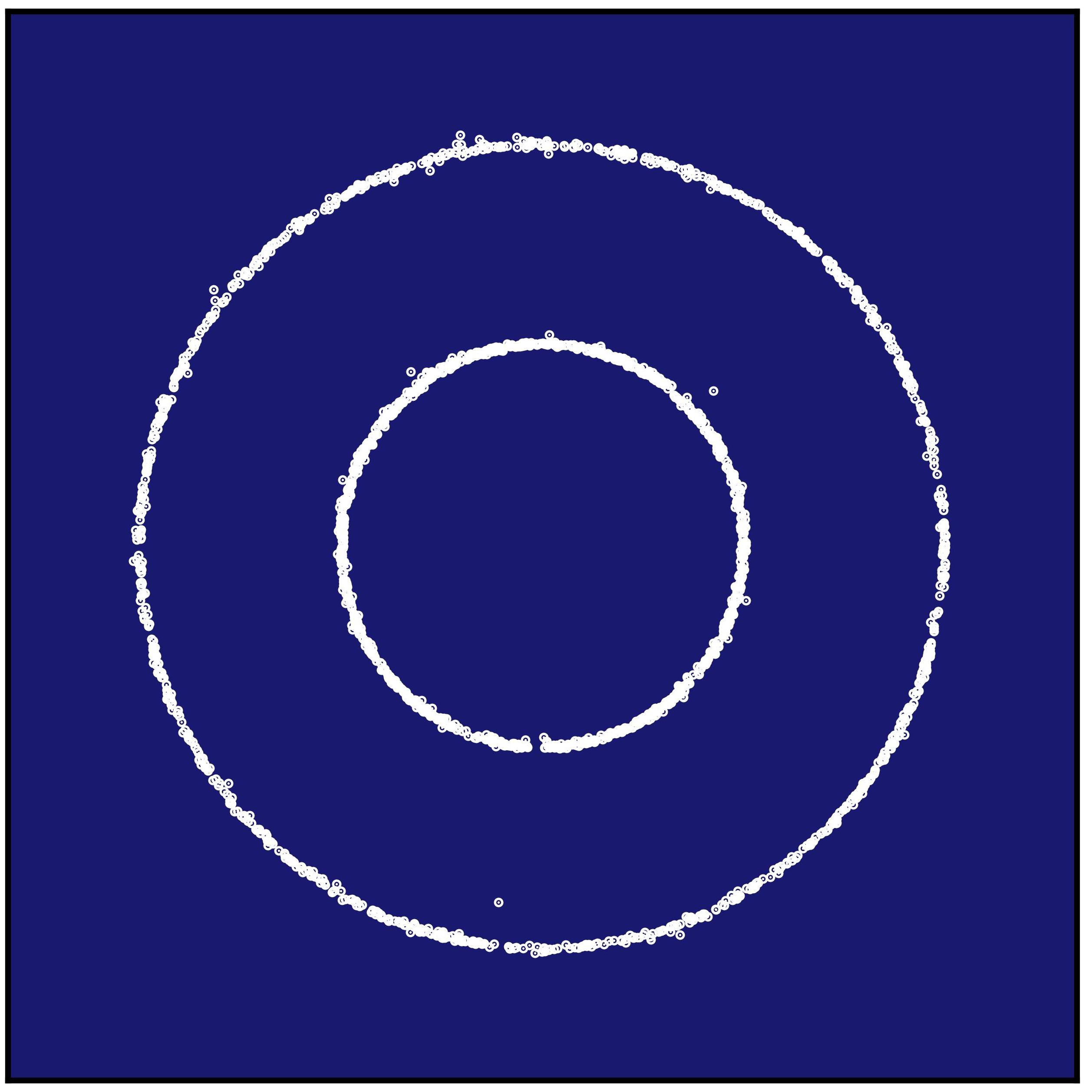}
        \includegraphics[width=0.48\textwidth]{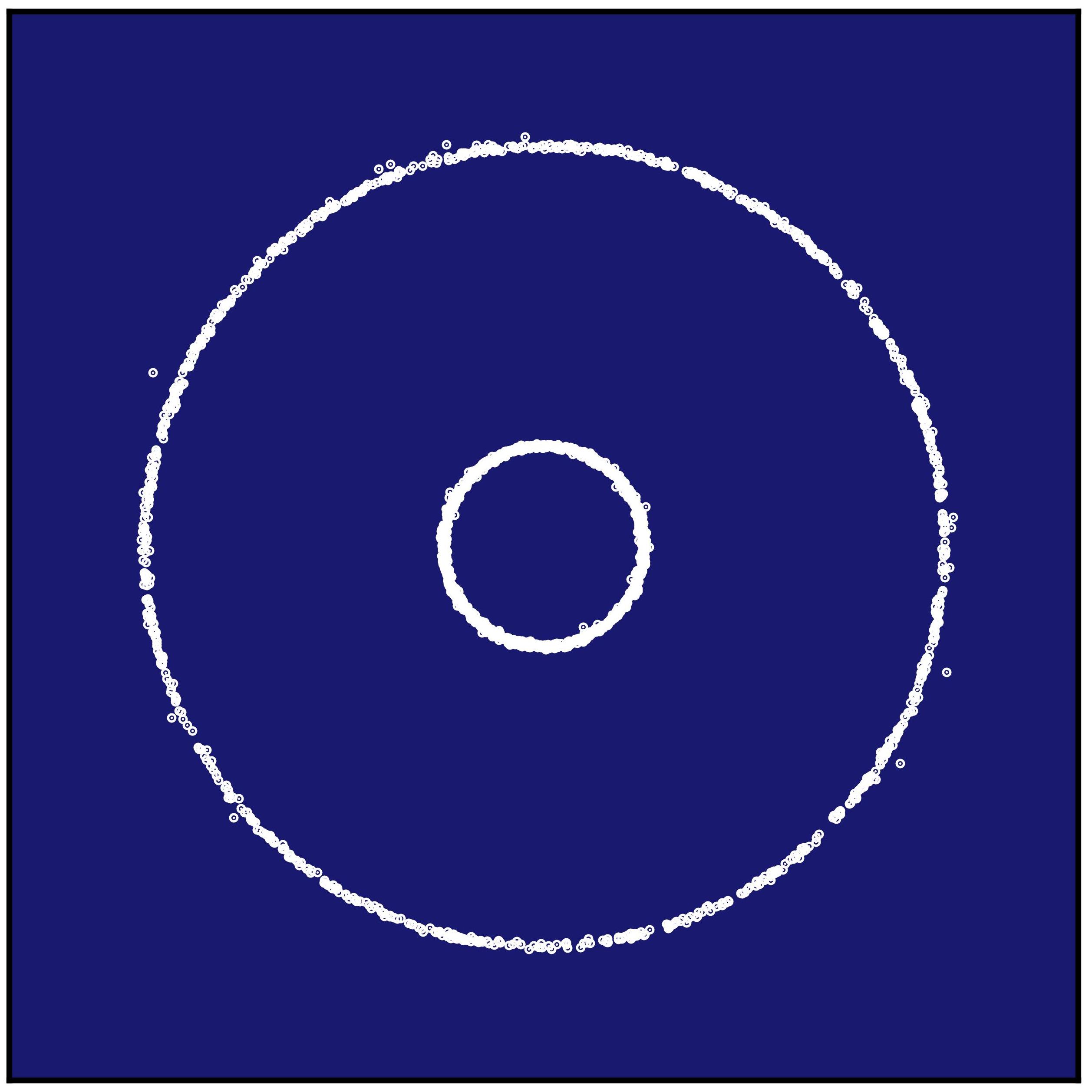}
        }
	\end{minipage}
\hspace{7pt}
    \begin{minipage}{0.13\linewidth}
        \vspace{4pt}
 		\centerline{\makecell{sines}}
		\vspace{10.5pt}
        \includegraphics[width=\textwidth]{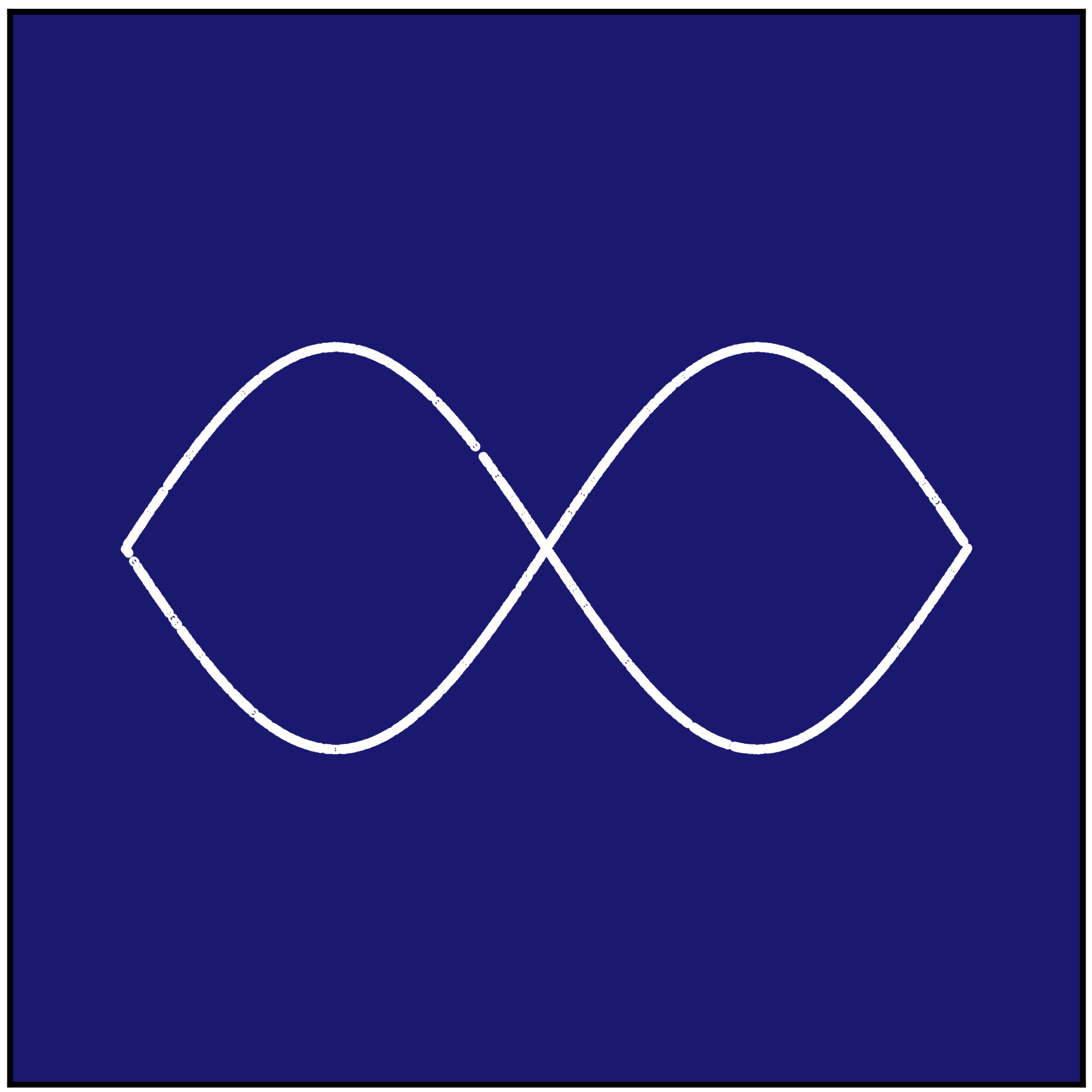}
        \vspace{2pt}
        \includegraphics[width=\textwidth]{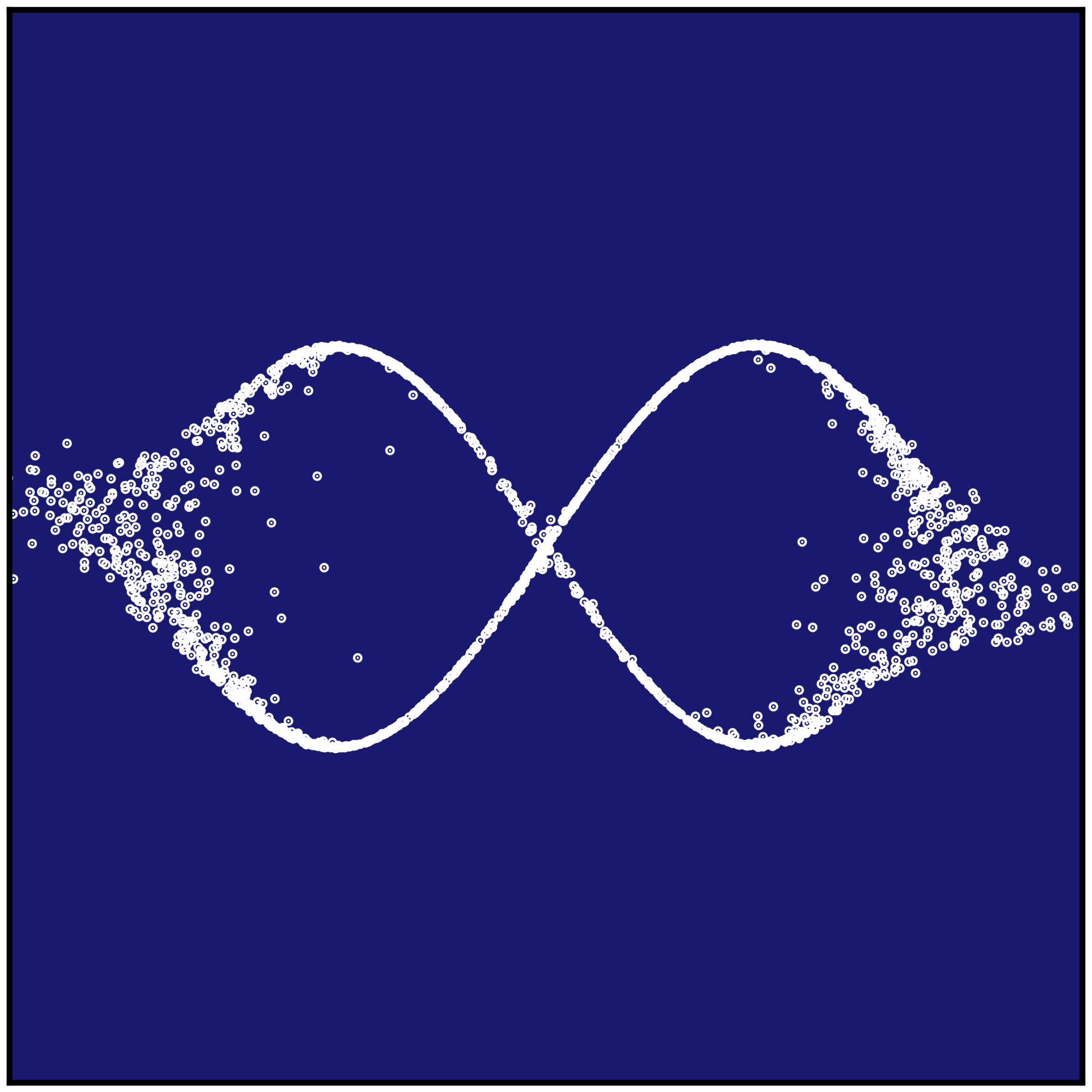}
        \vspace{2pt}
        \includegraphics[width=\textwidth]{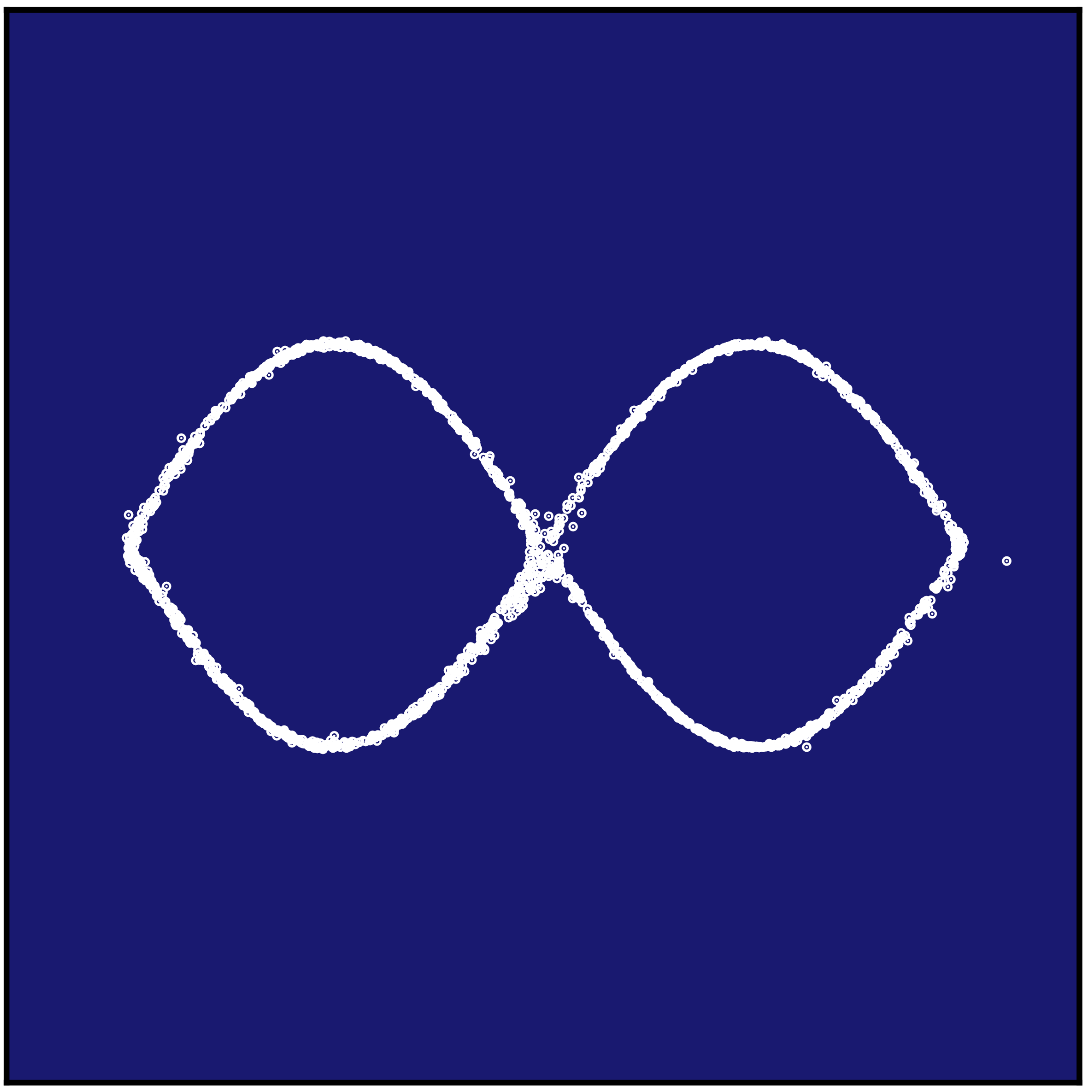}
        \vspace{2pt}
        \includegraphics[width=\textwidth]{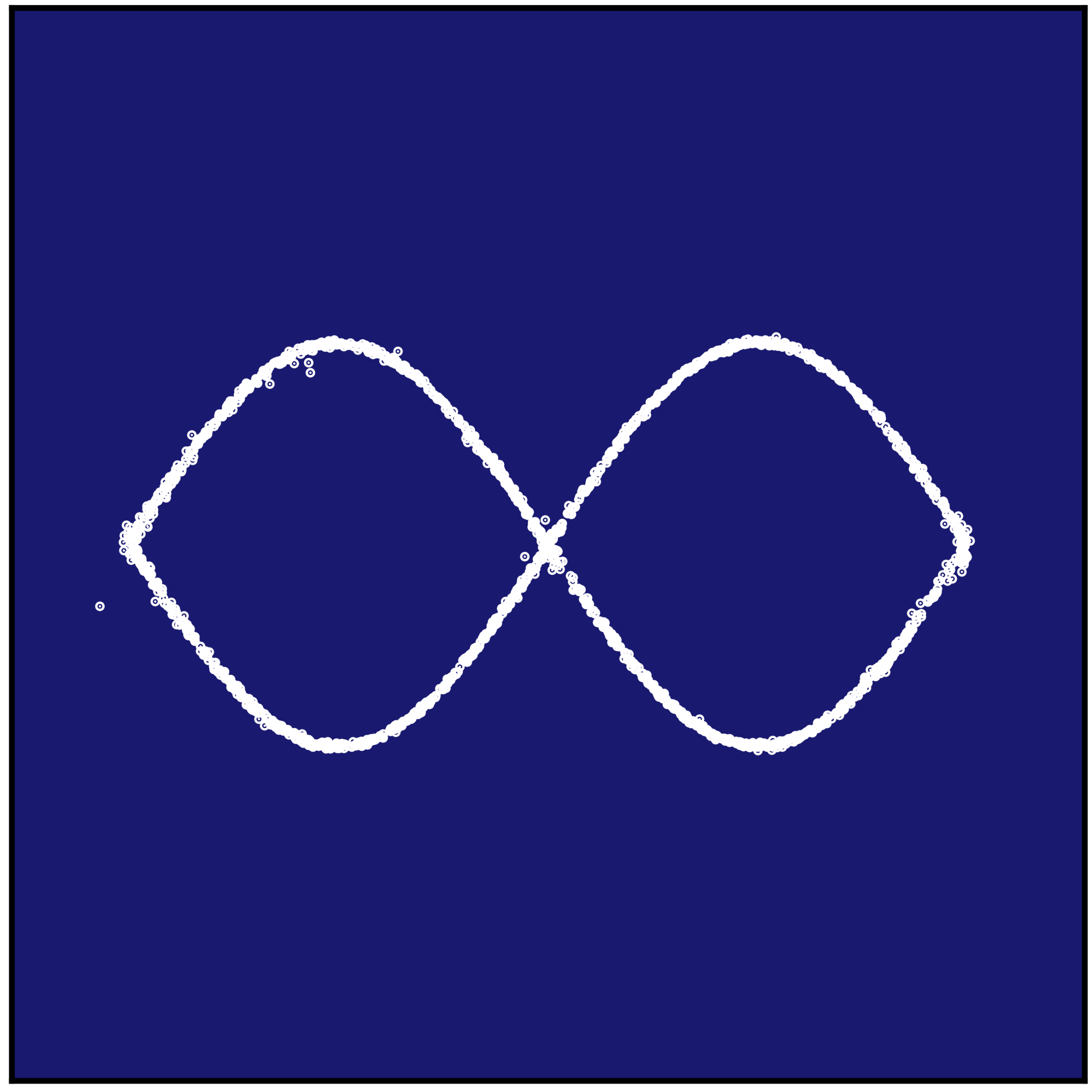}
	\end{minipage}
\hspace{2pt}
    \begin{minipage}{0.27\linewidth}
 		\centerline{\makecell{conditional sines\\\hspace{10pt}(c=0.6)\hspace{30pt}(c=0.3)}}
		\vspace{3pt}
        \subfloat{
        \includegraphics[width=0.48\textwidth]{figs/toy/sines_0.6.png}
        \includegraphics[width=0.48\textwidth]{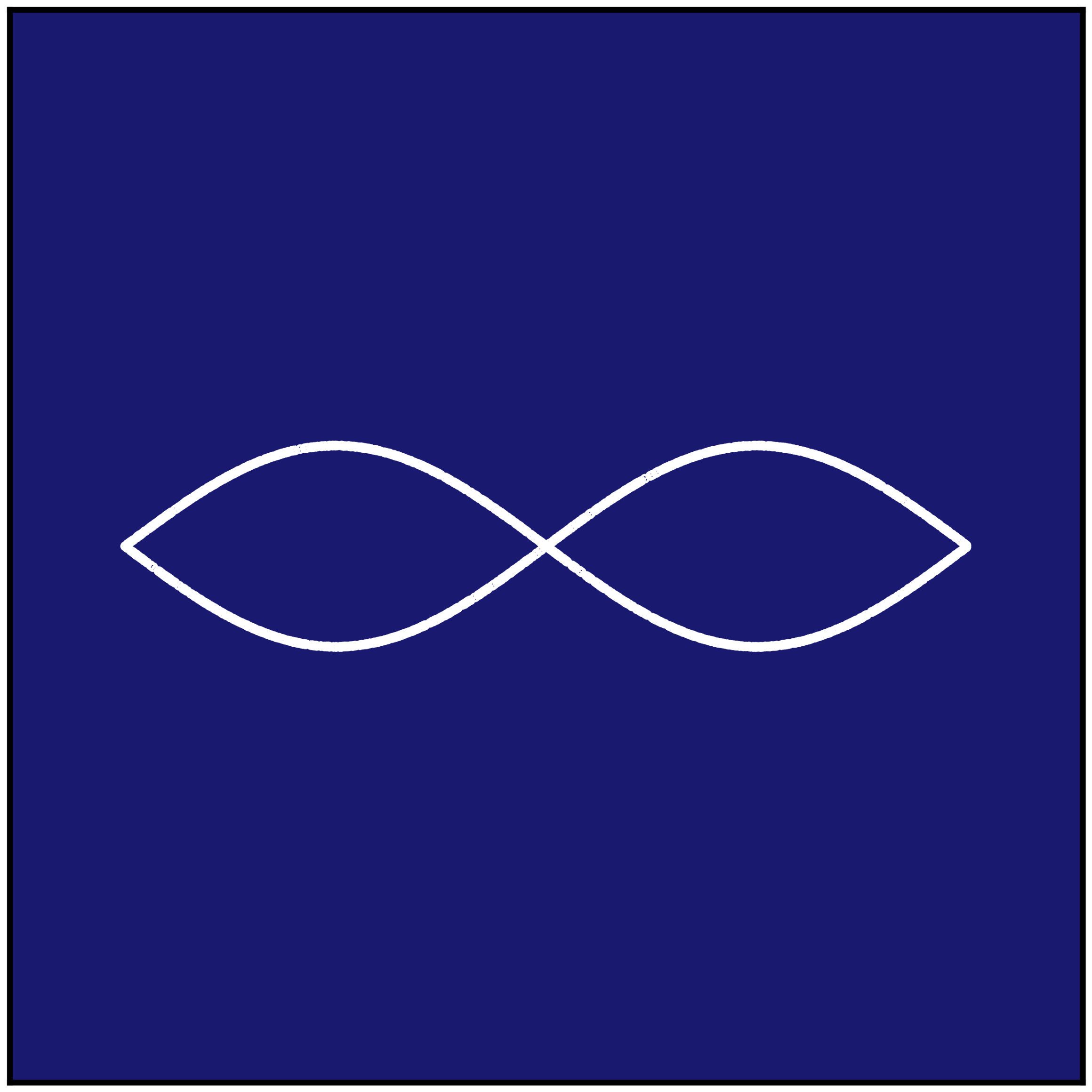}
        }\\[-8pt]
        \subfloat{
        \includegraphics[width=0.48\textwidth]{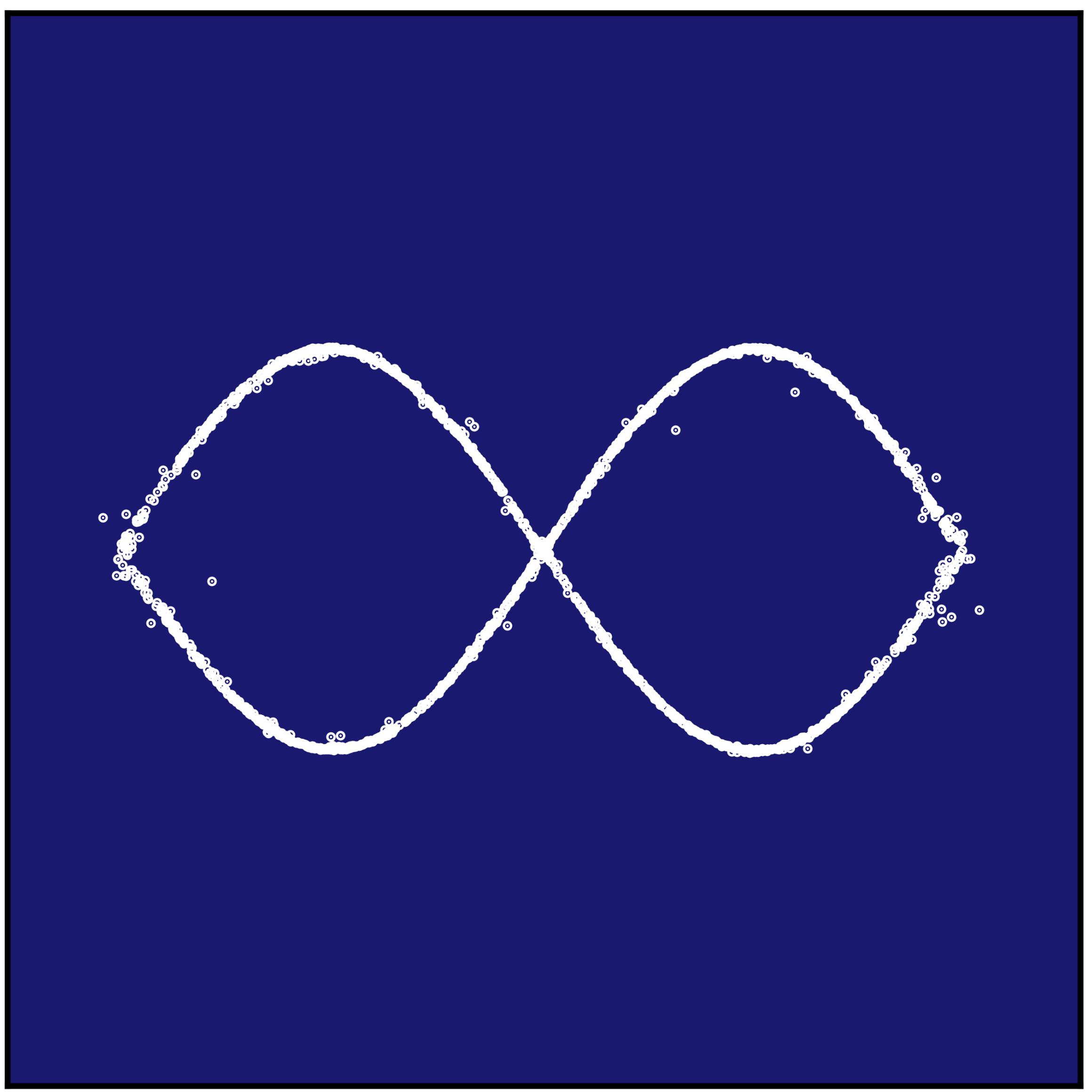}
        \includegraphics[width=0.48\textwidth]{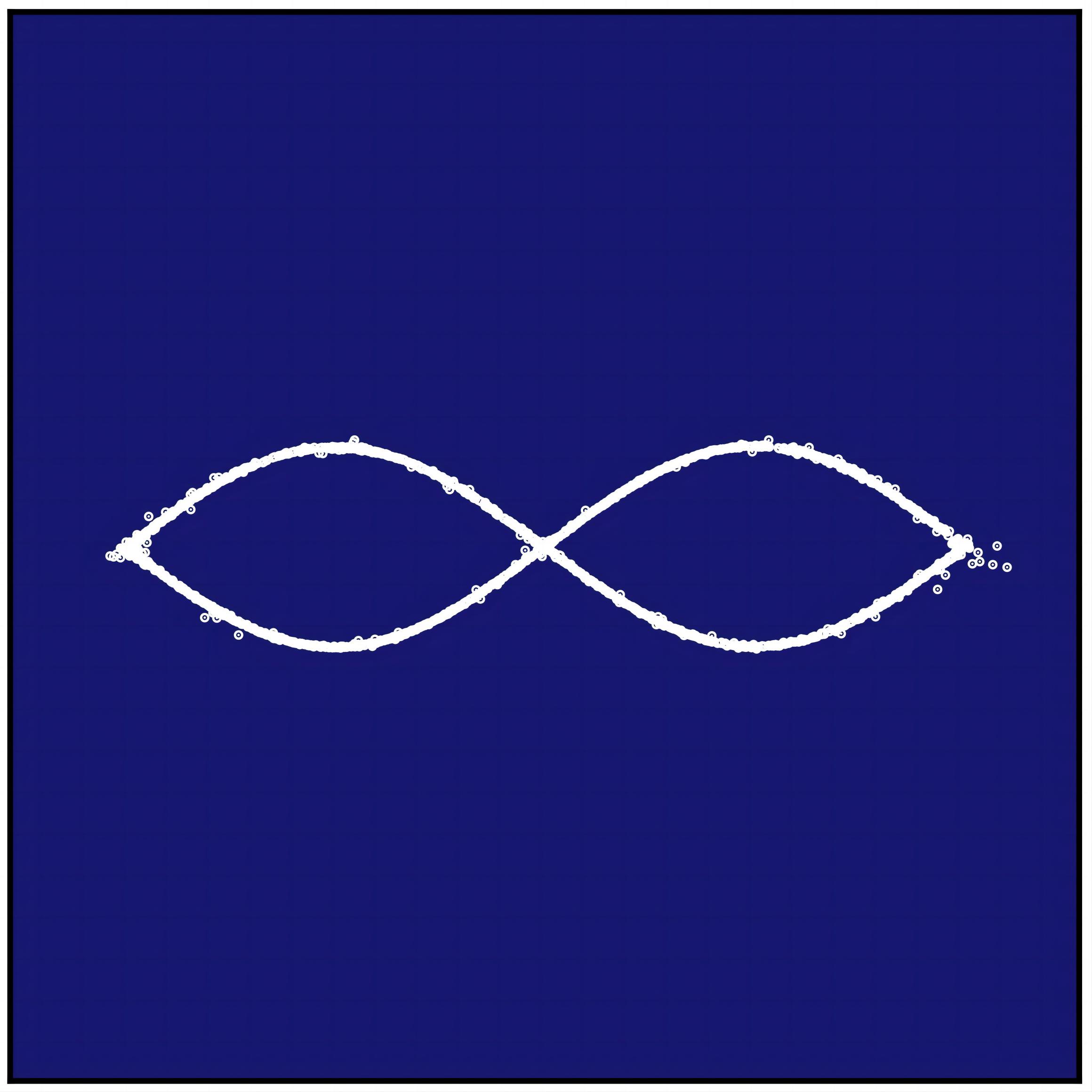}
        }\\[-8pt]
        \subfloat{
        \includegraphics[width=0.48\textwidth]{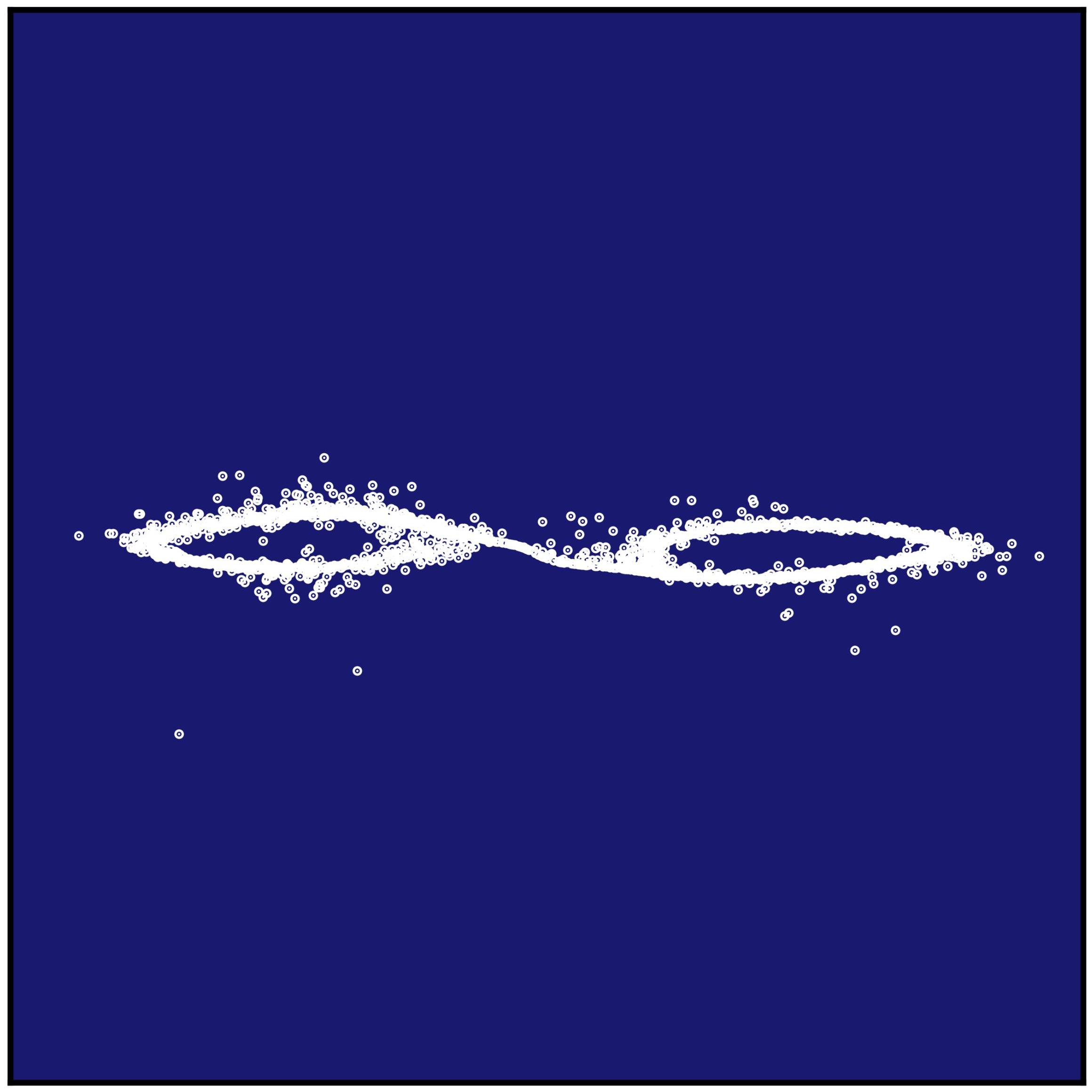}
        \includegraphics[width=0.48\textwidth]{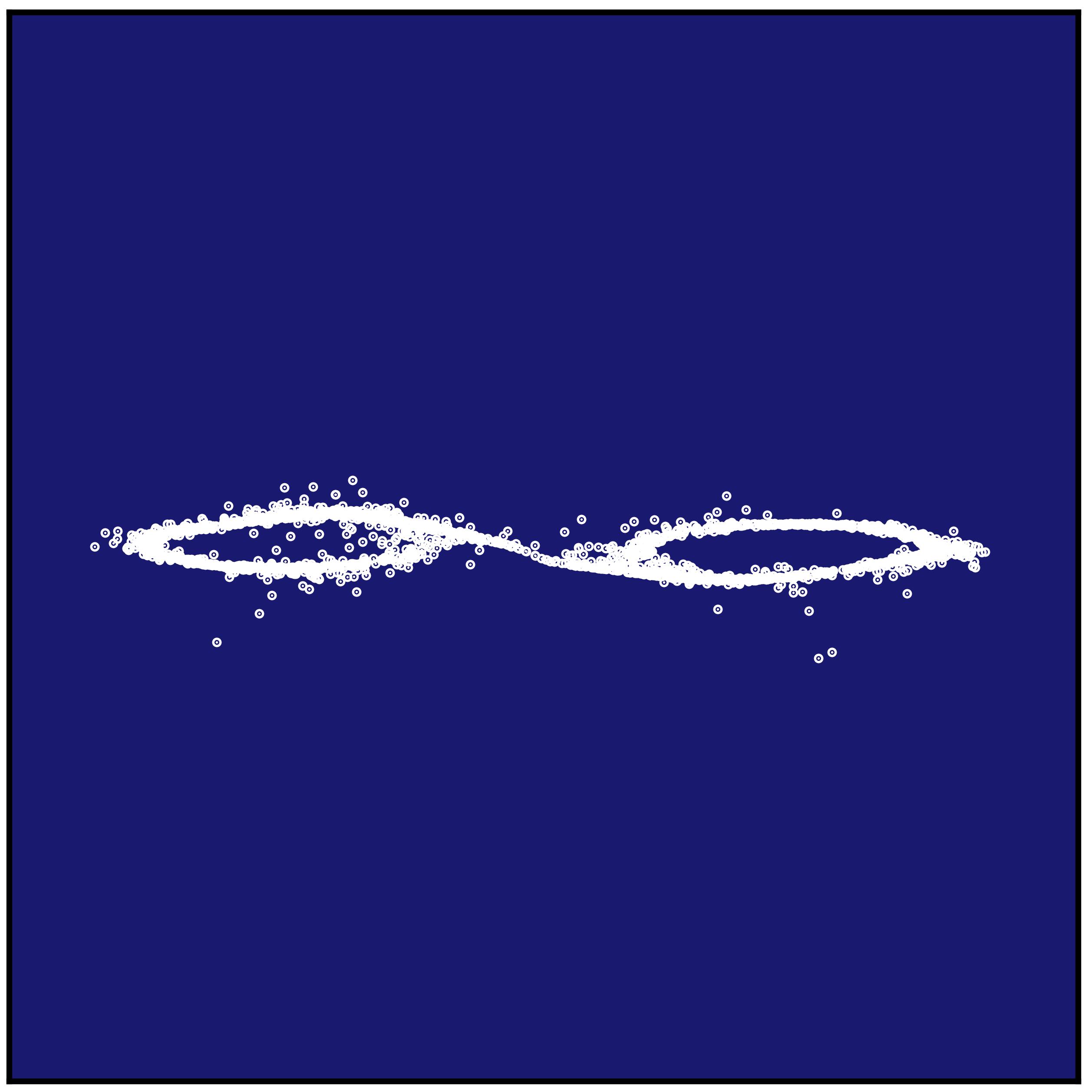}
        }\\[-8pt]
        \subfloat{
        \includegraphics[width=0.48\textwidth]{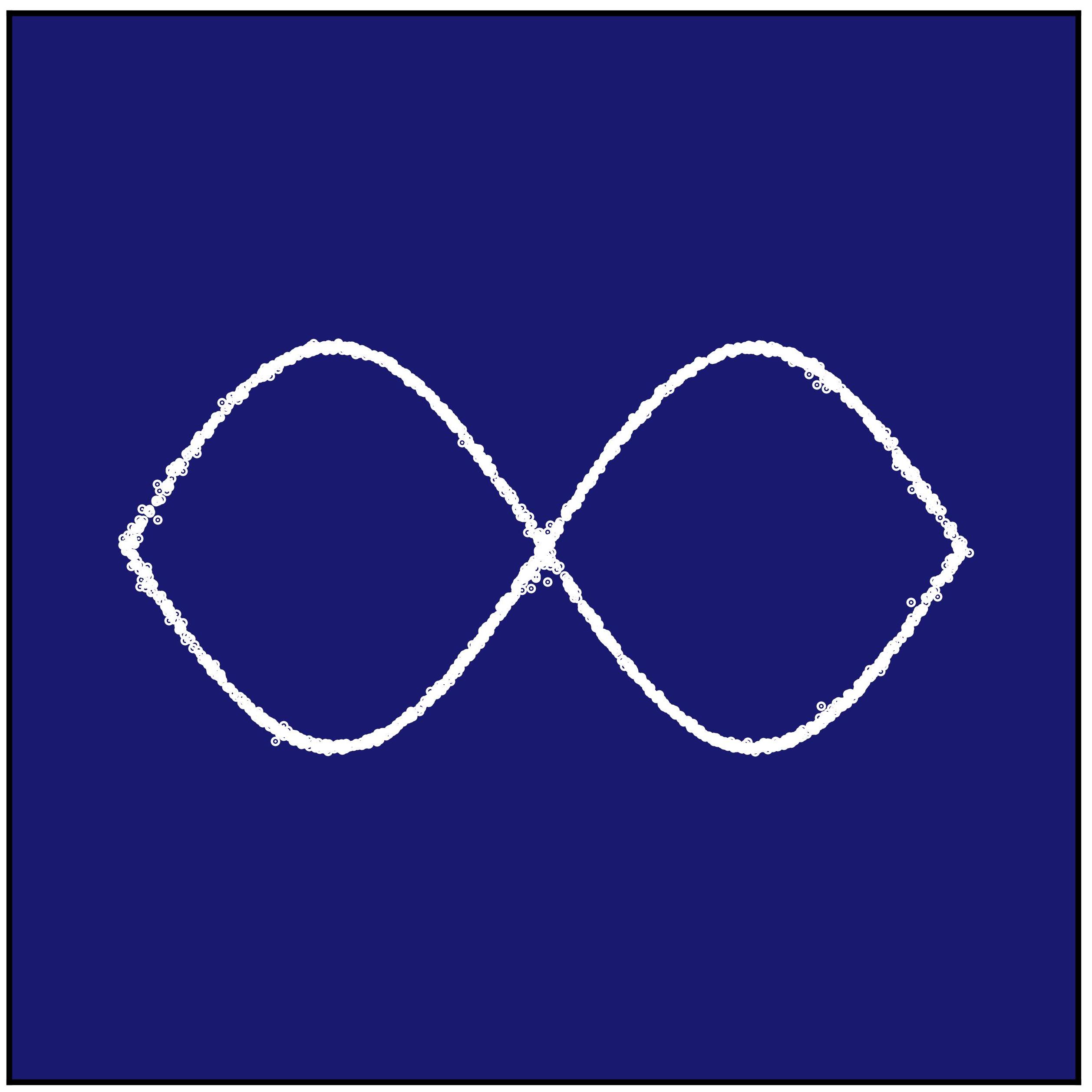}
        \includegraphics[width=0.48\textwidth]{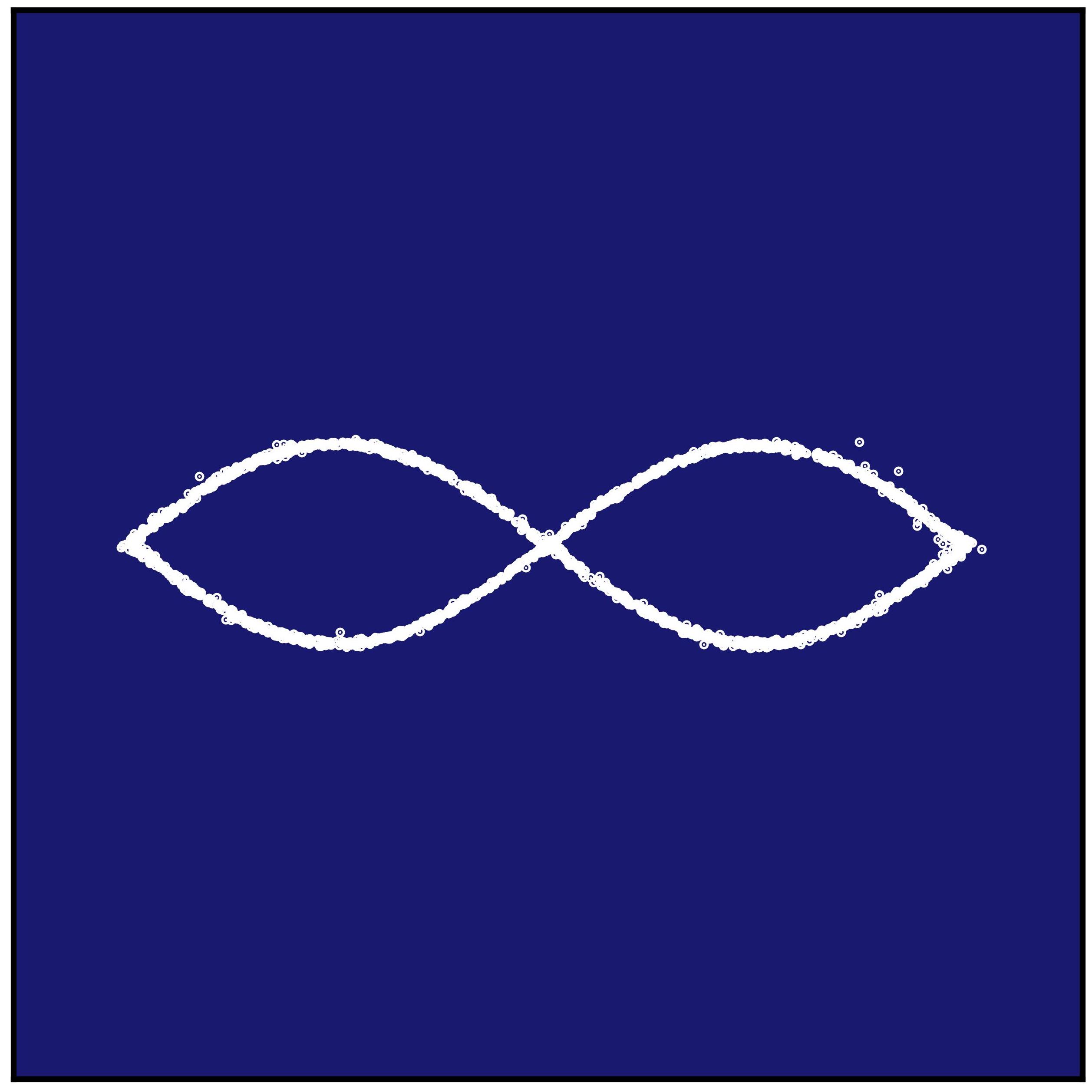}
        }
	\end{minipage}
    \caption{Comparison of FFJORD, SoftFlow, and PaddingFlow on 4 2-D distributions including 2 unconditional distributions (circles, and sines) and 2 conditional distributions (conditional circles, and conditional sines). }
    \label{fig.toy}
\end{figure*}

\subsection{PaddingFlow-based VAE Model}\label{sec:pd_vae}
The objective function of variational auto-encoder (VAE) is called evidence lower bound (ELBO)\cite{vae}, which can be written as:
\begin{equation}
\begin{aligned}
\mathcal{L}(\mathcal{I};\phi,\psi)=&\mathbb{E}_{q_{\phi}(z|\mathcal{I})}[-\mathrm{log}q_{\phi}(z|\mathcal{I})+\mathrm{log}p_{\psi}(\mathcal{I},z)].\label{eq.elbo}
\end{aligned}
\end{equation}
For flow-based VAE, the approximate posterior distribution $q_{\phi, \theta}(z|\mathcal{I})$ is obtained by transforming the initial distribution $q_{\phi}(x)$. The ELBO of flow-based VAE can be written as\cite{nfvae}:
\begin{equation}
\begin{aligned}
\mathcal{L}(\mathcal{I};\phi,\psi,\theta)
&=-\mathbb{E}_{q_{\phi}(x)}[\mathrm{log}q_{\phi}(x)-\mathrm{log|det}J_{F_{\theta}^{-1}}(x)|]\\
&+\mathbb{E}_{q_{\phi}(x)}[\mathrm{log}p_{\psi}(\mathcal{I},z)],\label{eq.vae_latent}
\end{aligned}
\end{equation}
where $x$ is sampling from $N(\mu,\Sigma_{\sigma})$, which is done by the reparameterization trick as:
\begin{equation}
\left\{
\begin{array}{ll}
(\mu,\sigma)=Q_{\phi}(\mathcal{I}) &\\[2pt]
x=\mu+\sigma\cdot\varepsilon, \mathrm{where}\ \varepsilon\sim N(0,I_{d})
\end{array}\right.\label{eq.repara}.
\end{equation}

\par To implement PaddingFlow on a VAE model, there are two main modifications, including implementing PaddingFlow noise and modifying the calculation of KL divergence. For the first one, we propose a more efficient way of computation:
\begin{itemize}
\vspace{5pt}
	\item\textbf{PaddingFlow Reparameterization} is another way of implementing PaddingFlow noise on a VAE model. Implementing PaddingFlow noise can be done by the method abovementioned (Eq. \ref{eq.add}) as shown in Fig. \ref{fig.vae} (green lines). The distribution of the data can be expressed as:
    \begin{equation}
    \left\{
    \begin{array}{ll}
    \mu'=(\mu,\vec{0}_{p}) &\\[4pt]
    \sigma'=((\sqrt{\sigma_{1}^{2}+a^{2}},\cdots,\sqrt{\sigma_{d}^{2}+a^{2}}),b\vec{1}_{p}) &\\[2pt]
    \end{array}\right.\label{eq.mu},
    \end{equation}
    However, for flow-based VAE models, due to variables of the distribution predicted by the encoder $Q_{\phi}$ being independent of each other, it can also be done by the PaddingFlow's version of reparameterization trick (Fig. \ref{fig.vae} blue lines):
    \begin{equation}
    \left\{
    \begin{array}{ll}
    (\mu,\sigma)=Q_{\phi}(\mathcal{I}) &\\[4pt]
    \mu'=(\mu,\vec{0}_{p}) &\\[4pt]
    \sigma'=(\sigma+a\vec{1}_{d},b\vec{1}_{p}) &\\[4pt]
    \varepsilon'\sim N(0,I_{d+p}) &\\[4pt]
    x'=\mu'+\sigma'\cdot\varepsilon'
    \end{array}\right.\label{eq.p_repara},
    \end{equation}
	where $\vec{1}_{p}$ denotes p-D 1-vector, and $\vec{0}_{p}$ denotes p-D 0-vector. The distribution of noise is different from adding noise directly under the same hyperparameter $a$, but choosing the different value of $a$ could obtain the same noise.
\vspace{5pt}
\end{itemize}
\par As for the loss function of VAE models, after adding padding-dimensional noise, the ELBO $\mathcal{L}$ should be written as:
\begin{equation}
\begin{aligned}
\mathcal{L}(\mathcal{I};\phi,\psi,\theta)
&=-\mathbb{E}_{q_{\phi}(x')}[\mathrm{log}q_{\phi}(x')]\\
&+\mathbb{E}_{q_{\phi}(x')}[\mathrm{log|det}J_{F_{\theta}^{-1}}(x')|]\\
&+\mathbb{E}_{q_{\phi}(x')}[\mathrm{log}p_{\psi}(\mathcal{I},z)].\label{eq.vae_loss}
\end{aligned}
\end{equation}
However, due to the distribution of noise being predetermined and padding dimensions being independent of data dimensions, the entropy of distribution $q_{\phi}(x')$ can be written as:
\begin{equation}
H[q_{\phi}(x')]=f(H[q_{\phi}(x)])+const,\label{eq.bijective}
\end{equation}
where $f(\cdot)$ is a nonlinear bijective function. Therefore, the second term of loss function $\mathcal{L}_{latent}$ can be simplified as follows:
\begin{equation}
\begin{aligned}
\mathcal{L}(\mathcal{I};\phi,\psi,\theta)
&=H[q_{\phi}(x)]\\
&+\mathbb{E}_{q_{\phi}(x')}[\mathrm{log|det}J_{F_{\theta}^{-1}}(x')|]\\
&+\mathbb{E}_{q_{\phi}(x')}[\mathrm{log}p_{\psi}(\mathcal{I},z)].\label{eq.p_vae_loss}
\end{aligned}
\end{equation}

\begin{figure}[t]
	\centering
	\subfloat{
		\includegraphics[width=0.45\textwidth]{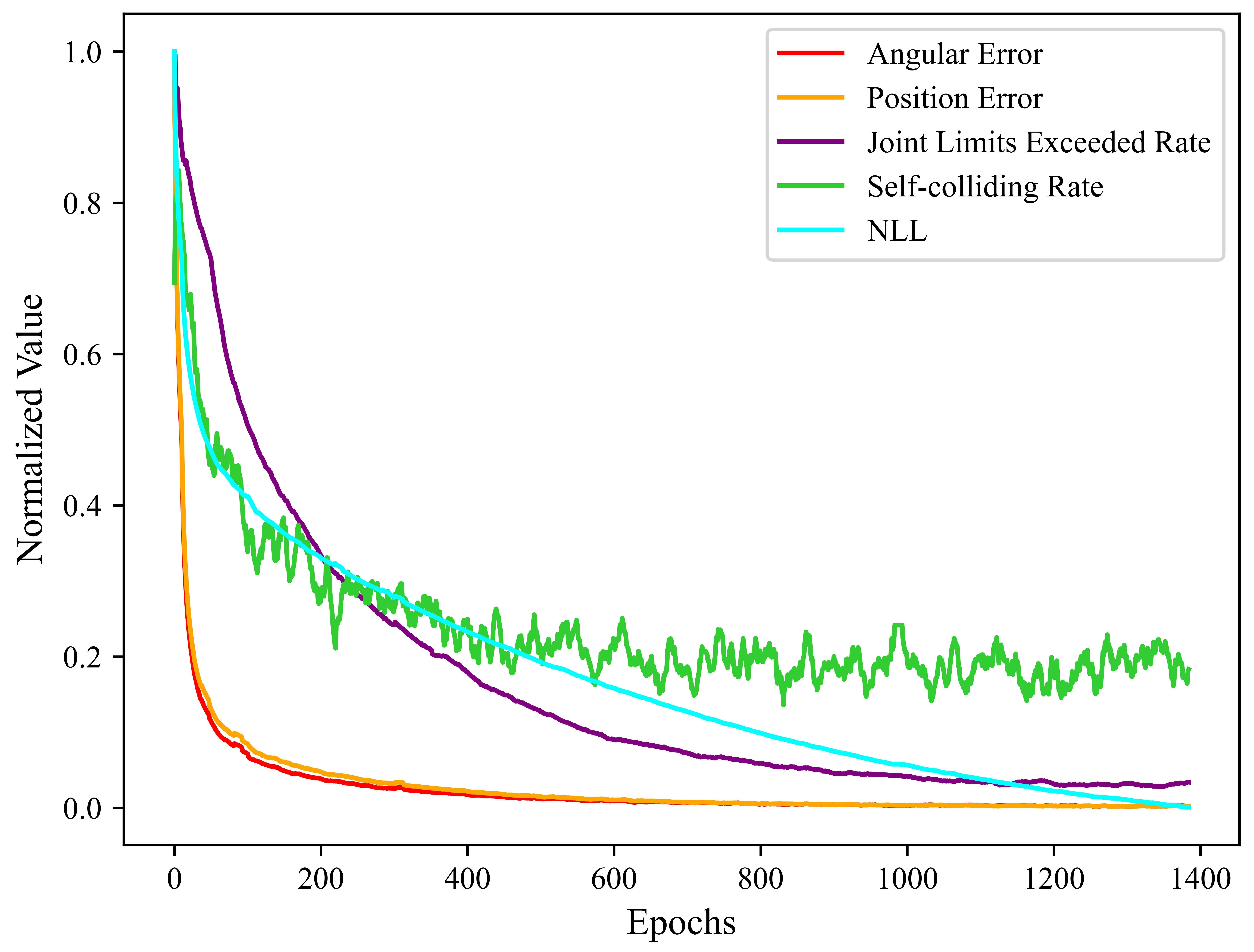}}
    \caption{Negative log-likelihood continuously decreases while all four metrics for evaluating IK solutions have been well-converged. }
    \label{fig.nll}
\end{figure}

\section{Experiments}\label{sec.exp}
In this section, firstly, we explain why existing metrics are not suitable for evaluating our method on density estimation tasks briefly, and introduce new metrics for evaluating density estimation models. Then, we evaluate PaddingFlow on unconditional density estimation (tabular datasets and VAE datasets), and conditional density estimation (IK experiments).

\subsection{Evaluation Metrics}\label{sec.exp_metrics}
Prior work uses log-likelihood to evaluate flow-based models on tabular datasets (Eq. \ref{eq.loss} and \ref{eq.cnf}). The trace and determinant of the Jacobian matrix should be calculated for computing log-likelihood. However, we didn't find a way to modify the ODE solver 
for computing the trace and the determinant without being affected by padding-dimensional noise. Besides, in the IK experiments, we observed negative log-likelihood continuously decreases while four metrics for evaluating IK solutions in robotics have been well-converged (shown in Fig. \ref{fig.nll}). Therefore, log-likelihood might not evaluate the quality of samples from flow-based models well. In this paper, we introduce several new metrics for evaluating density estimation.
\par To measure the similarity between the target distribution and the predicted distribution by two point sets ($X$, $Y$) sampled from such two distributions respectively, the used metrics should vary from different kinds of point sets. In this paper, we discuss which metrics are suitable for ordered point sets ($^{\mathcal{O}}X$, $^{\mathcal{O}}Y$) or disorder point sets ($^{\mathcal{D}}X$, $^{\mathcal{D}}Y$) respectively. We first describe three distance metrics, including Euclidean distance (L2), Chamfer distance (CD), and earth mover's distance (EMD). They can be defined as follows:

\begin{itemize}
\vspace{5pt}
	\item\textbf{Euclidean distance (L2):} To measure the distance between two ordered point sets ($^{\mathcal{O}}X$, $^{\mathcal{O}}Y$), L2 is the most common distance:
    \begin{equation}
    \mathrm{L2}(^{\mathcal{O}}X,^{\mathcal{O}}Y)=\sum_{i=1}^{N}\Vert ^{\mathcal{O}}x_{i}-{^{\mathcal{O}}y_{i}}\Vert _{2}.\label{eq.l2}
    \end{equation}
\vspace{5pt}
	\item\textbf{Chamfer distance (CD):} As for the distance between two disordered point sets ($^{\mathcal{D}}X$, $^{\mathcal{D}}Y$), CD assumes the two nearest points, each from the two given sets respectively, is the pair of corresponding points. Then the distance of such two points is used for computing the distance of the given point sets:
    \begin{equation}
    \begin{aligned}
    \mathrm{CD}(^{\mathcal{D}}X,^{\mathcal{D}}Y)=
    &\frac{1}{|^{\mathcal{D}}X|}\sum_{x\in ^{\mathcal{D}}X}\min_{y\in ^{\mathcal{D}}Y}D(x,y)\\
    &+\frac{1}{|^{\mathcal{D}}Y|}\sum_{y\in ^{\mathcal{D}}Y}\min_{x\in ^{\mathcal{D}}X}D(y,x),
    \end{aligned}
    \end{equation}
    where $D(\cdot,\cdot)$ is a distance measure. If the given sets have the same number of points and the distance measure is symmetric (2-norm), CD can be simplified as follows:
    \begin{equation}
    \mathrm{CD}(^{\mathcal{D}}X,^{\mathcal{D}}Y)=\sum_{x\in ^{\mathcal{D}}X}\min_{y\in ^{\mathcal{D}}Y}\Vert x-y\Vert _{2}^{2}.\label{eq.cd}
    \end{equation}
\vspace{5pt}
	\item\textbf{Earth mover's distance (EMD):} Instead of using a fixed distance measure, EMD is to fit a bijection for the given sets to find the pair of corresponding points:
    \begin{equation}
    \mathrm{EMD}(^{\mathcal{D}}X,^{\mathcal{D}}Y)=\min_{f:^{\mathcal{D}}X\rightarrow ^{\mathcal{D}}Y}\sum_{x\in ^{\mathcal{D}}X}\Vert x-f(x)\Vert _{2},\label{eq.emd}
    \end{equation}
	where $f(\cdot)$ is a bijective function between the point sets ($^{\mathcal{D}}X$, $^{\mathcal{D}}Y$).
\vspace{5pt}
\end{itemize}

\par The abovementioned distance metrics can evaluate a model, which fits on a single distribution (tabular datasets). However, in some tasks (experiments of VAE models), we need to evaluate a model that fits on multiple distributions. Therefore, we further describe two metrics that evaluate models by two sets of point sets, which are sampled from the target distributions and the predicted distributions ($S_t$, $S_p$):

\begin{itemize}
\vspace{5pt}
	\item\textbf{Minimum matching distance (MMD)} is the averaged distance between each point set in $S_t$ and its nearest neighbor in $S_p$:
    \begin{equation}
    \mathrm{MMD}(S_{t},S_{p})=\frac{1}{|S_{t}|}\sum_{X\in S_{t}}\min_{Y\in S_{P}}D(X,Y),\label{eq.mmd}
    \end{equation}
    where $D(\cdot,\cdot)$ can be L2, CD, or EMD. If there is only one point set ($X$) in $S_{t}$, it can be written as:
    \begin{equation}
    \mathrm{MMD}(X,S_{p})=\min_{Y\in S_{P}}D(X,Y),\label{eq.mmd_simp}
    \end{equation}
    which means it can also be used for evaluating models fitting on a single distribution.
\vspace{5pt}
	\item\textbf{Coverge (COV)} measures the rate of the point sets from predicted distribution in $S_p$ that can match the corresponding target distribution represented by a point set in $S_t$:
    \begin{equation}
    \hspace{-0.4cm} \mathrm{COV}(S_{t},S_{p})=\frac{|\{\arg \min_{Y\in S_p} D(X,Y)|X\in S_{t}\}|}{|S_{p}|},\label{eq.cov}
    \end{equation}
    where $D(\cdot,\cdot)$ can be L2, CD, or EMD.
\vspace{5pt}
\end{itemize}

\par In conclusion, for the experiments of tabular datasets, including UCI datasets and BSDS300, we use average CD, average EMD, MMD-CD, and MMD-EMD for evaluation. For the experiments of VAE models, we use MMD-L2 and COV-L2 for evaluation. As for IK experiments, we use two metrics for evaluating IK solutions in robotics, which are position error, and angular error.

\begin{table*}[t]
    \caption{Average CD, average EMD, MMD-CD, and MMD-EMD (lower is better) on the test set from UCI datasets and BSDS300. }
    \centering
    \setlength{\tabcolsep}{21.5pt}
    \begin{threeparttable}
    \begin{tabular}{clcccc}
        \toprule [1pt]\noalign{\vskip 2pt}
        \multirow{2}{*}[-2pt]{Dataset} & \multirow{2}{*}[-2pt]{Model} & \multirow{2}{*}[-2pt]{CD ($\downarrow$)} & \multirow{2}{*}[-2pt]{EMD ($\downarrow$)} 
        & \multicolumn{2}{c}{MMD ($\downarrow$)} \\[2pt]
        \cline{5-6}\noalign{\vskip 5pt}
        &&&& CD & EMD \\[2pt]
        \hline\noalign{\vskip 5pt}
        \multirow{3}{*}{\makecell{POWER \\ d=6; N=2,049,280}}
        &FFJORD                 & 0.153 & 0.116 & 0.144 & 0.111\\
        &PaddingFlow (1, 0)     & 0.145 &   0.107 & 0.137 & 0.101\\
        &PaddingFlow (1, 0.01)  & \textbf{0.142} & \textbf{0.105} & \textbf{0.135} & \textbf{0.0980} \\[5pt]
        \hline\noalign{\vskip 5pt}
        \multirow{3}{*}{\makecell{GAS \\ d=8; N=1,052,065}}
        &FFJORD                 & 1.29 & 0.146 & 0.950 & 0.135\\
        &PaddingFlow (1, 0)     & 1.18 & \textbf{0.131} & 0.913& \textbf{0.121}\\
        &PaddingFlow (3, 0)     & \textbf{0.890} & 0.141 & \textbf{0.390} & 0.128 \\[5pt]
        \hline\noalign{\vskip 5pt}
        \multirow{2}{*}{\makecell{HEPMASS \\ d=21; N=525,123}}
        &FFJORD                 & 13.8 & 0.164 & 13.8 & 0.158 \\
        &PaddingFlow (1, 0)     & \textbf{13.8} & \textbf{0.161} & \textbf{13.7} & \textbf{0.153}\\[5pt]
        \hline\noalign{\vskip 5pt}
        \multirow{5}{*}{\makecell{MINIBOONE \\ d=43; N=36,488}}
        &FFJORD                 & 24.6 & 0.270 & 24.1 & \textbf{0.254}\\
        &PaddingFlow (1, 0)     & 24.7 & 0.269 & 24.4 & 0.256\\
        &PaddingFlow (1, 0.01)  & \textbf{24.5} & \textbf{0.268} & 24.1 & 0.255\\
        &PaddingFlow (2, 0)     & 24.6 & 0.270 & 24.2 & 0.255 \\
        &PaddingFlow (2, 0.01)  & 24.6 & 0.271 & \textbf{24.0} & 0.257 \\[5pt]
        \hline\noalign{\vskip 5pt}
        \multirow{3}{*}{\makecell{BSDS300 \\ d=63; N=1,300,000}}
        &FFJORD                 & 0.683 &   0.0281 & 0.548 & 0.0227\\
        &PaddingFlow (10, 0)    & 0.592 &   0.0255 & 0.484 & \textbf{0.0212} \\
        &PaddingFlow (10, 0.01) & \textbf{0.495} & \textbf{0.0248} & \textbf{0.480} & 0.0218 \\[2pt]
        \bottomrule [1pt]
    \end{tabular}
    \vspace{5pt}
    \begin{tablenotes}
    \footnotesize
    \item[*]The hyperparameters p, and a shown in the names of models denoted as PaddingFlow (p, a) in the tabular represent the number of padding dimensions and the variance of data noise respectively. The variance of the padding-dimensional noise is set to 2.
    \end{tablenotes}
    \end{threeparttable}
\label{tab.tabular}
\end{table*}

\subsection{Density Estimation on 2D Artificial Data}\label{sec.toy}
We designed four 2-D artificial distributions to visually exhibit the performance of FFJORD, SoftFlow, and PaddingFlow on both unconditional and conditional distributions. In this section, models of SoftFlow and PaddingFlow are both based on the same model of FFJORD. SoftFlow noise variance $c$ is sampled from $U(0,0.1)$. The hyperparameters of PaddingFlow $p$, $a$, and $b$ are set to 1, 0.01, and 2 respectively. 
\par As shown in Fig. \ref{fig.toy}, FFJORD can not fit into a manifold well, except conditional sines. After implementing SoftFlow noise, the performance on two unconditional distributions of FFJORD is well-improved, but degraded significantly on two conditional distributions at the same time. In contrast, PaddingFlow can perform well on both unconditional and conditional distributions.

\begin{figure*}[t]
\hspace{-5pt}
  	\begin{minipage}{0.05\linewidth}
 		\centerline{\footnotesize \rotatebox[origin=c]{90}{Data}}
        \vspace{4pt}
        \centerline{\footnotesize \rotatebox[origin=c]{90}{FFJORD}}
        \vspace{4pt}
        \centerline{\footnotesize \rotatebox[origin=c]{90}{Ours}}
	\end{minipage}
\hspace{-8pt}
 	\begin{minipage}{0.45\linewidth}
        \includegraphics[width=\textwidth]{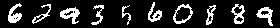}
        \includegraphics[width=\textwidth]{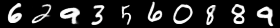}
        \includegraphics[width=\textwidth]{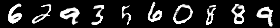}
	\end{minipage}
\hspace{8pt}
 	\begin{minipage}{0.45\linewidth}
        \includegraphics[width=\textwidth]{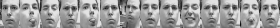}
        \includegraphics[width=\textwidth]{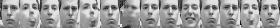}
        \includegraphics[width=\textwidth]{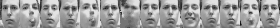}
	\end{minipage}
    \caption{Comparison of VAE models based on FFJORD, and PaddingFlow on MNIST and Frey Faces. }
    \label{fig.vae_c}
\end{figure*}

\begin{table*}[t]
    \caption{Cross entropy (lower is better), MMD (lower is better), and COV (higher is better) on the test set for VAE models.}
    \centering
    \setlength{\tabcolsep}{20.5pt}
    \begin{threeparttable}
    \begin{tabular}{clcccc}
        \toprule [1pt]\noalign{\vskip 2pt}
        & Model & MNIST & Omniglot & Frey Faces & Caltech\\[5pt]
        \hline\noalign{\vskip 5pt}
        \multirow{2}{*}{Cross Entropy ($\downarrow$)}
        &FFJORD           & 55.9 & 64.3 & 1715.6  & 63.0 \\
        &PaddingFlow      & \textbf{36.1} & \textbf{64.0} & \textbf{1666.9} & \textbf{60.2} \\[5pt]
        \hline\noalign{\vskip 5pt}
        \multirow{2}{*}{MMD ($\downarrow$)}
        &FFJORD           & 17.3 & 20.5 & 0.834 & 18.6 \\
        &PaddingFlow      & \textbf{11.0} & \textbf{20.3} & \textbf{0.621} & \textbf{17.9} \\[5pt]
        \hline\noalign{\vskip 5pt}
        \multirow{2}{*}{COV (\%, $\uparrow$)}
        &FFJORD           & 96.4 & \textbf{99.0} &  100.0 & \textbf{98.8} \\
        &PaddingFlow      & \textbf{100.0} & 98.8 & \textbf{100.0} & 98.7 \\[2pt]
        \bottomrule [1pt]
    \end{tabular}
    \vspace{5pt}
    \begin{tablenotes}
    \footnotesize
    \item[*]PaddingFlow noise in VAE experiments is only padding-dimensional noise.
    \vspace{5pt}
    \item[**]The distance measure used in the experiments of VAE models is L2.
    \end{tablenotes}
    \end{threeparttable}
\label{tab.vae}
\end{table*}

\begin{table}[!hb]
    \caption{The results on the test set (150,000 random IK problems of Panda manipulator).}
    \centering
    \begin{threeparttable}
    \begin{tabular}{lccccc}
        \toprule [1pt]\noalign{\vskip 2pt}
        Model & Position Error (mm, $\downarrow$) & Angular Error (deg, $\downarrow$)\\[5pt]
        \hline\noalign{\vskip 5pt}
        GLOW                 & 7.37 & 0.984 \\
        SoftFlow (IKFlow)    & 6.86 & 2.382 \\[5pt]
        \hline\noalign{\vskip 5pt}
        PaddingFlow          & \textbf{5.96} & \textbf{0.621} \\[2pt]
        \bottomrule [1pt]
    \end{tabular}
    \end{threeparttable}
\label{tab.ik}
\end{table}

\begin{figure}[!hb]
	\centering
	\subfloat{
		\includegraphics[height=0.2\textwidth]{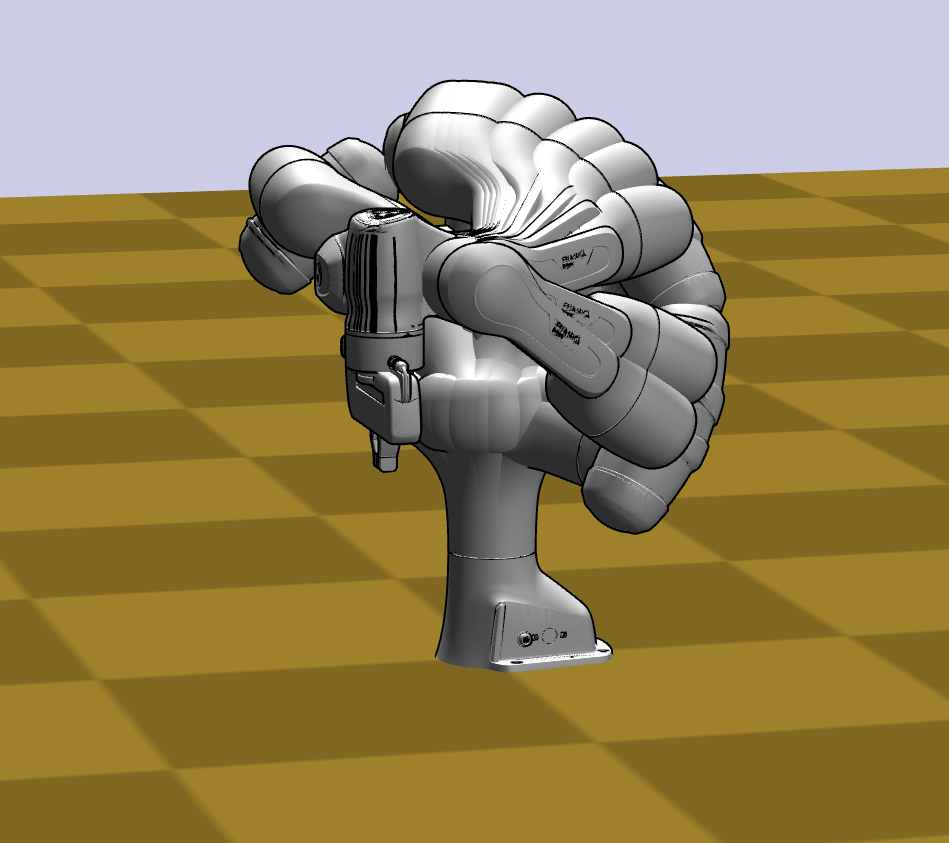}
        \includegraphics[height=0.2\textwidth]{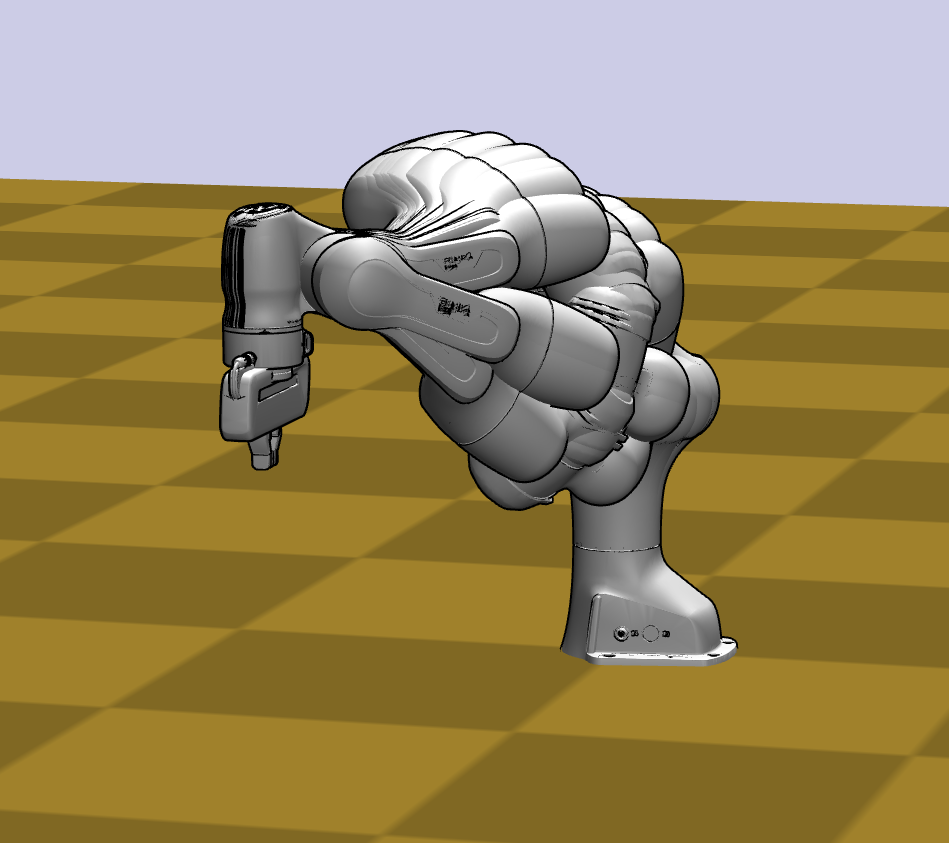}}\\[3pt]
    \subfloat{
		\includegraphics[height=0.2\textwidth]{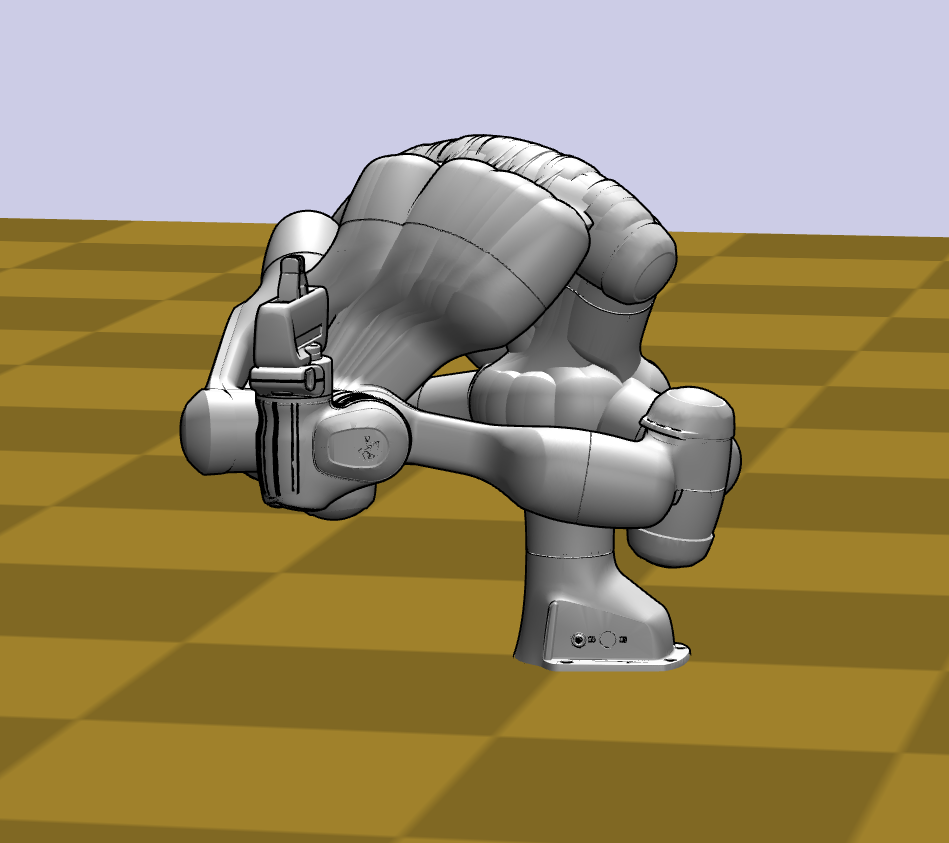}
        \includegraphics[height=0.2\textwidth]{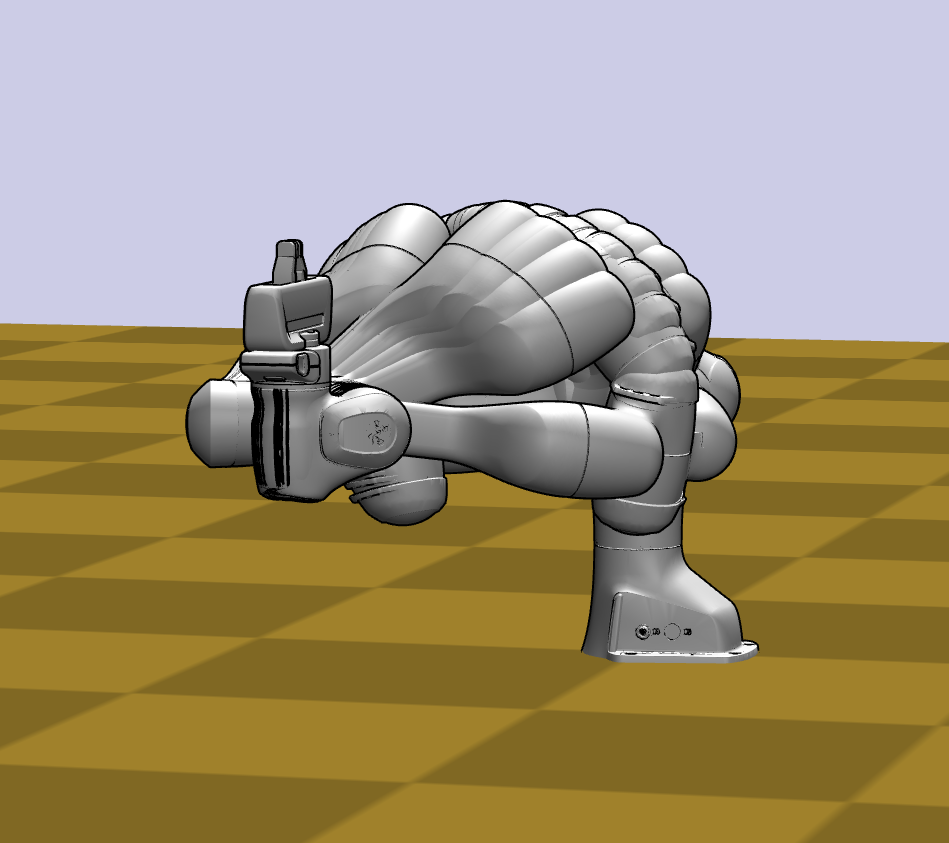}}
    \caption{IK solutions for a Panda manipulator reaching to given end effector poses. }
    \label{fig.ik-solutions}
\end{figure}

\subsection{Unconditional Density Estimation}\label{sec.exp_unc}
In this section, we compare FFJORD and PaddingFlow on tabular datasets and VAE models. The results show our method can improve normalizing flows on the main benchmarks of density estimation, including five tabular datasets and four image datasets for VAE models.

\vspace{10pt}
\subsubsection{Tabular Datasets}\label{sec.exp_tab}
We evaluate unconditional density estimation on five tabular datasets, including four UCI datasets, and BSDS300, which are preprocessed as \cite{maf}. In this section, our method is implemented on FFJORD. Average CD, average EMD, MMD-CD, and MMD-EMD are used for comparison of FFJORD and PaddingFlow. The results (Tab. \ref{tab.tabular}) show that PaddingFlow performs better than FFJORD across all five tabular datasets. Especially on the GAS and BSDS300 datasets, the improvement of implementing our method is significant. On the POWER, GAS, HEPMASS, and BSDS300 datasets, we observed improvement in the experiments that only concatenate padding-dimensional noise on the data. Additionally, throughout the tabular experiments, we found it difficult to find a suitable number of padding dimensions $p$ for individual datasets (i.e. BSDS300).
\par In general, our method can improve the performance of normalizing flows on tabular datasets, but the selection of the hyperparameter $p$ is intractable sometimes.

\vspace{10pt}
\subsubsection{Variational Autoencoder}\label{sec.exp_vae}
We also evaluate unconditional density estimation in variational inference on four image datasets, including MNIST, Omniglot, Frey Faces, and Caltech 101 Silhouettes. All four datasets are preprocessed as \cite{Sylvester}. In this section, our method is implemented on VAE models using FFJORD. Cross Entropy, MMD-L2, and COV-L2 are used for comparison of FFJORD and PaddingFlow. The results (Tab. \ref{tab.vae}) show that PaddingFlow performs better than FFJORD. Especially for the MNIST and Frey Faces, the improvement is significant, and the images reconstructed by PaddingFlow-based VAE models have higher quality as shown in Fig. \ref{fig.vae_c}, which are with richer details. Moreover, the best models of the four datasets are all without data noise, and the dimension of padding-dimensional noise is relatively low. 
\par In general, for flow-based VAE models, PaddingFlow can bring improvement via only padding-dimensional noise with relatively low dimension, which means the selection of hyperparameters is quite simple.

\subsection{Conditional Density Estimation}\label{sec.exp_c}
Inverse Kinematics (IK) is to map the work space to the robot's joint space. IK is an important prior task for many tasks in robotics, such as motion planning. IK solvers are required to return a set of IK solutions that covers the universal set of solutions, as shown in Fig. \ref{fig.ik-solutions}, in case of no feasible path or only suboptimal paths. Due to such a feature, efficient sampling of normalizing flows is suitable for IK. In this section, we choose two metrics in robotics for evaluating IK solutions, including position error, and angular error. As for two other metrics mentioned in Fig. \ref{fig.nll} which are joint limits exceeded rate and self-colliding rate, due to the feature of IK abovementioned, they are just the complement of position error and angular error. Furthermore, both SoftFlow and PaddingFlow provide no improvement to joint limits exceeded rate and self-colliding rate. Thus, such two metrics are not used in this section. In IK experiments, SoftFlow and PaddingFlow are based on the same model of GLOW. The architecture of GLOW follows models used in experiments of Panda manipulator in \cite{ikflow}. PaddingFlow noise in IK experiments is only padding-dimensional noise. The hyperparameters of padding-dimensional noise $p$, and $b$ are set to 1, and 2 respectively. The variance of SoftFlow noise is sampled from $U(0,0.001)$.
\par As shown in Tab. \ref{tab.ik}, PaddingFlow performs better than both GLOW and SoftFlow on the position error and angular error.

\section{Conclusion}\label{sec.clusion}
In this paper, we propose PaddingFlow, a novel dequantization method, which satisfies the five key features of an ideal dequantization method we list (Sec. \ref{sec.5charac}). In particular, PaddingFlow overcomes the limitation of existing dequantization methods, which is that they have to change the data distribution. We validate our method on the main benchmarks of unconditional density estimation, and conditional density estimation. The results show our method performs better in all experiments and can improve both discrete normalizing flows and continuous normalizing flows.

\newpage
\begin{IEEEbiography}[{\includegraphics[width=1in,height=1.25in,clip,keepaspectratio]{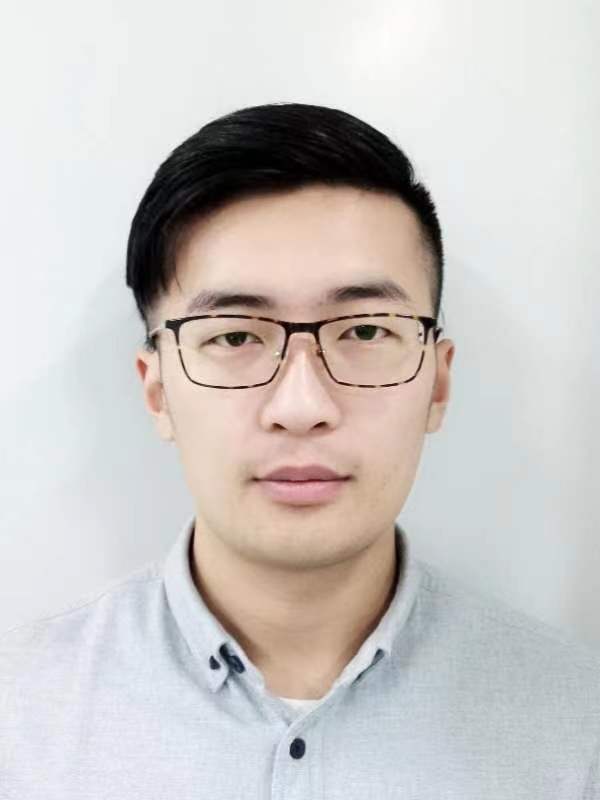}}]{Qinglong Meng}
received the B.E. degree in measurement and control technology and instrumentation from the College of Instrumentation and Electrical Engineering, Jilin University, Changchun, China, in 2017, He is currently working toward the M.S. degree in electronic information with Tsinghua Shenzhen International Graduate School, Tsinghua University, Shenzhen, China.
\par His research interests include robotics, motion planning, and machine learning in robotics.
\end{IEEEbiography}

\vspace{-340pt}
\begin{IEEEbiography}[{\includegraphics[width=1in,height=1.25in,clip,keepaspectratio]{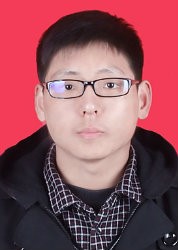}}]{Chongkun Xia}
received the Ph.D. degree in pattern recognition and intelligent systems from Northeastern University, Shenyang, China, in 2021.
\par He is an associate professor in the School of Advanced Manufacturing, Sun Yat-sen University. His research interests include robotics, motion planning and machine learning.
\end{IEEEbiography}

\vspace{-330pt}
\begin{IEEEbiography}[{\includegraphics[width=1in,height=1.25in,clip,keepaspectratio]{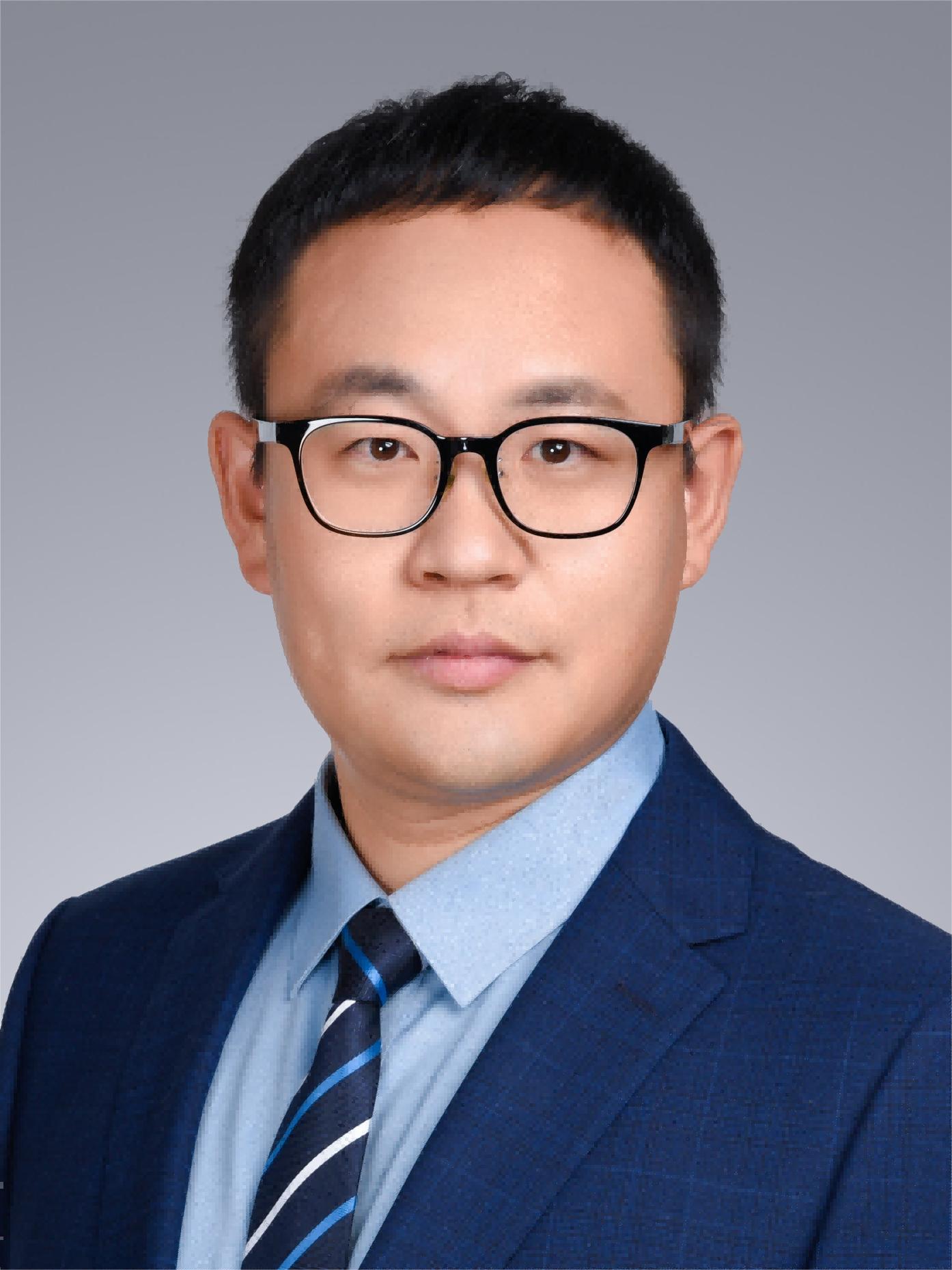}}]{Xueqian Wang}
received the B.E. degree in mechanical design, manufacturing, and automation from the Harbin University of Science and
Technology, Harbin, China, in 2003, and the M.Sc. degree in mechatronic engineering and the Ph.D. degree in control science and engineering from the Harbin Institute of Technology (HIT), Harbin, China, in 2005 and 2010, respectively.
\par From June 2010 to February 2014, he was the Post-doctoral Research Fellow with HIT. He is currently a Professor and the Leader of the Center of Intelligent Control and Telescience, Tsinghua Shenzhen International Graduate School, Tsinghua University, Shenzhen, China. His research interests include dynamics modeling, control, and teleoperation of robotic systems.
\end{IEEEbiography}

\newpage
\section*{APPENDIX}

\begin{figure*}[t]
\hspace{15pt}
  	\begin{minipage}{0.05\linewidth}
        \vspace{0.6\linewidth}
 		\centerline{\small\rotatebox[origin=c]{90}{MNIST}}
        \vspace{2.8\linewidth}
        \centerline{\small\rotatebox[origin=c]{90}{Omniglot}}
        \vspace{2.8\linewidth}
        \centerline{\small\rotatebox[origin=c]{90}{Frey Faces}}
        \vspace{2.8\linewidth}
        \centerline{\small\rotatebox[origin=c]{90}{Caltech}}
	\end{minipage}
\hspace{-10pt}
 	\begin{minipage}{0.4\linewidth}
        \vspace{4pt}
 		\centerline{Data}
        \vspace{5pt}
        \includegraphics[width=\textwidth]{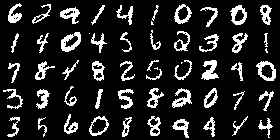}
        \vspace{2pt}
        \includegraphics[width=\textwidth]{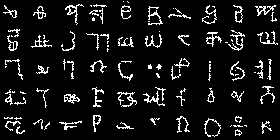}
        \vspace{2pt}
        \includegraphics[width=\textwidth]{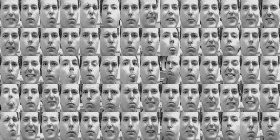}
        \vspace{2pt}
        \includegraphics[width=\textwidth]{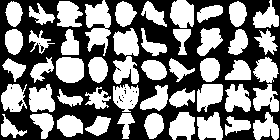}
	\end{minipage}
\hspace{2pt}
     \begin{minipage}{0.4\linewidth}
        \vspace{4pt}
 		\centerline{Reconstructions}
        \vspace{5pt}
        \includegraphics[width=\textwidth]{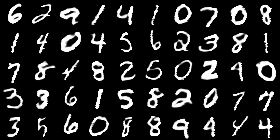}
        \vspace{2pt}
        \includegraphics[width=\textwidth]{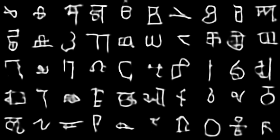}
        \vspace{2pt}
        \includegraphics[width=\textwidth]{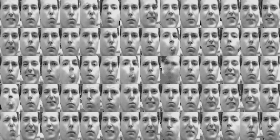}
        \vspace{2pt}
        \includegraphics[width=\textwidth]{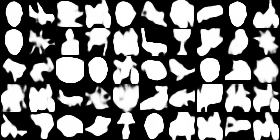}
	\end{minipage}
    \caption{Data, and reconstructions from PaddingFlow-based VAE models trained on MNIST, Omniglot, Frey Faces, and Caltech 101 Silhouettes respectively. }
    \label{fig.recon}
\end{figure*}

\vspace{10pt}

\subsection{Reconstructions from PaddingFlow-base VAE models}\label{sec.app_samples}
The pictures shown in Fig. \ref{fig.recon} are reconstructed from PaddingFlow-based VAE models trained on MNIST, Omniglot, Frey Faces, and Caltech 101 Silhouettes respectively.

\subsection{Proof of Eq. \ref{eq.exp_uni}}\label{sec.app_proof}
Here, the details of estimating the expectation of $Y$ is shown:
\begin{equation}
\begin{aligned}
\mathbb{E}(Y)&=\int_{-\infty}^{+\infty}ydy\int_{0}^{1}\frac{1}{\sqrt{2\pi}}e^{-\frac{1}{2}(y-u)^{2}}du\\
&=\int_{-\infty}^{+\infty}dy\int_{0}^{1}\frac{1}{\sqrt{2\pi}}(y-u)e^{-\frac{1}{2}(y-u)^{2}}du\\
&+\int_{-\infty}^{+\infty}dy\int_{0}^{1}\frac{1}{\sqrt{2\pi}}ue^{-\frac{1}{2}(y-u)^{2}}du\\
&\triangleq \mathcal{I}_{1}+\mathcal{I}_{2}.\label{eq.2int}
\end{aligned}
\end{equation}
The first integral ($\mathcal{I}_{1}$) can be easily computed:
\begin{equation}
\begin{aligned}
\mathcal{I}_{1}&=\int_{-\infty}^{+\infty}\frac{1}{\sqrt{2\pi}}\left( e^{-\frac{1}{2}(y-1)^{2}}-e^{-\frac{1}{2}y^{2}}\right )dy\\
&=0.
\end{aligned}
\end{equation}
The second integral ($\mathcal{I}_{2}$) can be further transformed as follows:
\begin{equation}
\begin{aligned}
\mathcal{I}_{2}&=\int_{-\infty}^{+\infty}dy\int_{0}^{1}\frac{1}{\sqrt{2\pi}}ue^{-\frac{1}{2}(y-u)^{2}}du\\
&=\int_{-\infty}^{+\infty}\frac{1}{\sqrt{2\pi}}e^{-\frac{1}{2}y^{2}}dy\int_{0}^{1}ue^{-\frac{1}{2}u^{2}}e^{yu}du.
\end{aligned}
\end{equation}
Due to $u\in [0,1]$, The term ($e^{yu}$) can be bounded as follows:
\begin{equation}
1\leqslant e^{yu}=(e^{u})^{y}\leqslant e^{y}.
\end{equation}
Therefore, the lower bound of $\mathcal{I}_{2}$ is:
\begin{equation}
\begin{aligned}
\mathcal{I}_{2}&\geqslant \int_{-\infty}^{+\infty}\frac{1}{\sqrt{2\pi}}e^{-\frac{1}{2}y^{2}}dy\int_{0}^{1}ue^{-\frac{1}{2}u^{2}}du\\
&=1-e^{-\frac{1}{2}}.
\end{aligned}
\end{equation}
The upper bound of $\mathcal{I}_{2}$ is:
\begin{equation}
\begin{aligned}
\mathcal{I}_{2}&\leqslant \int_{-\infty}^{+\infty}\frac{1}{\sqrt{2\pi}}e^{-\frac{1}{2}y^{2}+y}dy\int_{0}^{1}ue^{-\frac{1}{2}u^{2}}du\\
&=1-e^{-\frac{1}{2}}.
\end{aligned}
\end{equation}
In conclusion, according to the Squeeze Theorem:
\begin{equation}
\mathbb{E}(Y)=1-e^{-\frac{1}{2}}.
\end{equation}

\subsection{Experimental details}\label{sec.app_exp}
On the tabular datasets and VAE experiments, we follow the settings in \cite{ffjord} over network architectures. The hyperparameters of models used in Sec. \ref{sec.exp_tab} can be found in Tab. \ref{tab.tabular_settings}. The hyperparameters of VAE models used in Sec. \ref{sec.exp_vae} can be found in Tab. \ref{tab.vae_settings}. 

\begin{table*}[t]
    \caption{Hyperparameters for models on tabular datasets.}
    \centering
    \begin{threeparttable}
    \begin{tabular}{lccccc}
        \toprule [1pt]\noalign{\vskip 2pt}
        Dataset & nonlinearity & \# layers & hidden dim multiplier & \# flow steps & batch size\\[5pt]
        \hline\noalign{\vskip 5pt}
        HEPMASS     & softplus & 2 & 10 & 10 & 10000 \\
        Others      & softplus & 2 & 20  & 1 & 1000 \\[5pt]
        \bottomrule [1pt]
    \end{tabular}
    \vspace{5pt}
    \end{threeparttable}
\label{tab.tabular_settings}
\end{table*}

\begin{table*}[t]
    \caption{Hyperparameters for VAE models.}
    \centering
    \begin{threeparttable}
    \begin{tabular}{lcccccc}
        \toprule [1pt]\noalign{\vskip 2pt}
        Dataset & nonlinearity & \# layers & hidden dimension & \# flow steps & batch size & padding dimension\\[5pt]
        \hline\noalign{\vskip 5pt}
        MNIST       & softplus & 2 & 1024 & 2 & 64 & 2\\
        Omniglot    & softplus & 2 & 512  & 5 & 20 & 2\\
        Frey Faces  & softplus & 2 & 512  & 2 & 20 & 3\\
        Caltech     & tanh     & 1 & 2048 & 1 & 20 & 2\\[5pt]
        \bottomrule [1pt]
    \end{tabular}
    \vspace{5pt}
    \begin{tablenotes}
    \footnotesize
    \item[*]The variances of data noise and padding-dimensional noise are set to 0, and 2 respectively.
    \end{tablenotes}
    \end{threeparttable}
\label{tab.vae_settings}
\end{table*}


\begin{thebibliography}{1}
\bibitem{softflow}
H. Kim, H. Lee, W. H. Kang, J. Y. Lee, and N. S. Kim, “SoftFlow: Probabilistic Framework for Normalizing Flow on Manifolds,” in Advances in Neural Information Processing Systems, 2020.
\bibitem{discrete}
B. Uria, I. Murray, and H. Larochelle, “RNADE: The real-valued neural autoregressive density-estimator,” in Advances in Neural Information Processing Systems, 2013.
\bibitem{flowpp}
J. Ho, X. Chen, A. Srinivas, Y. Duan, and P. Abbeel, “Flow++: Improving Flow-Based Generative Models with Variational Dequantization and Architecture Design,” in Proceedings of the 36th International Conference on Machine Learning, 2019.
\bibitem{survey}
G. Papamakarios, E. Nalisnick, D. J. Rezende, S. Mohamed, and B. Lakshminarayanan, “Normalizing Flows for Probabilistic Modeling and Inference,” in Journal of Machine Learning Research, vol. 22, no. 57, pp. 1-64, 2021.
\bibitem{cnf}
R. T. Q. Chen, Y. Rubanova, J. Bettencourt, and D. Duvenaud, “Neural ordinary differential equations,” in Advances in Neural Information Processing Systems, 2018.
\bibitem{ffjord}
W. Grathwohl, R. T. Q. Chen, J. Bettencourt, I. Sutskever, and D Duvenaud, "FFJORD: Free-form Continuous Dynamics for Scalable Reversible Generative Models," in arXiv preprint arXiv:1810.01367, 2018.
\bibitem{mnist}
Y. LeCun, C. Cortes, and C. J.C. Burges, "The MNIST database of handwritten digits," in Neural Computation, 10(5), 1191-1232.
\bibitem{uci}
J. N. V. Rijn and J. K. Vis, "Endgame Analysis of Dou Shou Qi," in arXiv preprint arXiv:1604.07312, 2014.
\bibitem{vae}
D. P. Kingma, and M. Welling, “Auto-Encoding Variational Bayes,” in arXiv preprint arXiv:1312.6114, 2022.
\bibitem{nfvae}
D. J. Rezende, and S. Mohamed, “Variational Inference with Normalizing Flows,” in Proceedings of the 32nd International Conference on Machine Learning, 2015.
\bibitem{maf}
G. Papamakarios, T. Pavlakou, and I. Murray, “Masked Autoregressive Flow for Density Estimation,” in arXiv preprint arXiv:1705.07057, 2017.
\bibitem{Sylvester}
R. V. D. Berg, L. Hasenclever, J. M. Tomczak, and M. Welling, “Sylvester Normalizing Flows for Variational Inference,” in arXiv preprint  	arXiv:1803.05649, 2018.
\bibitem{ikflow}
B. Ames, J. Morgan, and G. Konidaris, "IKFlow: Generating Diverse Inverse Kinematics Solutions," in IEEE Robotics and Automation Letters, vol. 7, no. 3, pp. 7177-7184, July 2022, doi: 10.1109/LRA.2022.3181374.
\bibitem{metrics}
P. Achlioptas, O. Diamanti, I. Mitliagkas, and L. Guibas, “Learning representations and generative models for 3d point clouds,” in Proceedings of the 35th International Conference on Machine Learning, 2018.

\end{thebibliography}
\end{document}